%% file: main.tex
\pdfoutput=1
\documentclass[11pt]{article}

\usepackage[T1]{fontenc}
\usepackage[utf8]{inputenc}
\usepackage{mathptmx}
\usepackage[scaled=0.92]{helvet}
\IfFileExists{t1lmtt.fd}{}{}
\usepackage[letterpaper,margin=1in]{geometry}
\usepackage{amsmath,amssymb,amsthm}
\usepackage{microtype}
\usepackage{graphicx}
\usepackage{booktabs}
\usepackage{tabularx}
\usepackage{multirow}
\usepackage{makecell}
\usepackage{enumitem}
\usepackage[table]{xcolor}
\usepackage{pifont}
\usepackage{caption}
\usepackage{subcaption}
\usepackage{placeins}
\usepackage{tikz}
\usetikzlibrary{arrows.meta,positioning,calc,fit,backgrounds,shapes.geometric}
\usepackage{fancyhdr}
\usepackage{orcidlink}
\usepackage[round,sort&compress]{natbib}
\usepackage{hyperref}
\usepackage[capitalize,nameinlink,noabbrev]{cleveref}

\input{macros}
\input{results-macros}

\input{v2-macros}
\input{v3-macros}
\hypersetup{colorlinks=true,linkcolor=linkblue,citecolor=linkblue,urlcolor=linkblue,
  pdftitle={AskWearables: Computed, Minimum-Necessary Context for AI Agents over Personal Wearable Data},
  pdfauthor={Sai Varun Kodathala, Prashanth Pollishetty, Jaylen Cargill}}

\title{\bfseries AskWearables\\[2pt]
\Large Computed, Minimum-Necessary Context for AI Agents over Personal Wearable Data}
\author{%
  Sai Varun Kodathala\,\orcidlink{0009-0002-3124-633X}\\
  SportsVision AI\\
  Minnetonka, MN\\
  \href{mailto:varun@sportsvision.ai}{\texttt{varun@sportsvision.ai}}
  \and
  Prashanth Pollishetty\\
  SportsVision AI\\
  Minnetonka, MN\\
  \href{mailto:prashanth@sportsvision.ai}{\texttt{prashanth@sportsvision.ai}}
  \and
  Jaylen Cargill\\
  SportsVision AI\\
  Minnetonka, MN\\
  \href{mailto:jaylen@sportsvision.ai}{\texttt{jaylen@sportsvision.ai}}}
\date{}

\begin{document}
\maketitle
\thispagestyle{fancy}

\input{sections/abstract}
\input{sections/introduction}
\input{sections/related}
\input{sections/design}
\input{sections/study}
\input{sections/results}

\input{sections/discussion}
\input{sections/scope}
\input{sections/conclusion}
\input{sections/ethics}

\bibliographystyle{unsrtnat}
\bibliography{references}

\clearpage
\appendix
\input{sections/appendix}

\end{document}

%% file: macros.tex
\definecolor{linkblue}{HTML}{1F5FAD}
\definecolor{aok}{HTML}{DCEFE3}
\definecolor{abad}{HTML}{F8D8CC}
\definecolor{ana}{HTML}{F1F0EC}
\definecolor{condA}{HTML}{2A78D6}   
\definecolor{condB}{HTML}{EB6834}   
\definecolor{condC}{HTML}{1BAF7A}   
\definecolor{inkgray}{HTML}{52514E}
\definecolor{pendingred}{HTML}{C0262D}

\newcommand{\sys}{AskWearables}
\newcommand{\code}[1]{\texttt{#1}}
\newcommand{\RQ}[1]{\textbf{RQ#1}}
\newcommand{\ok}{\ding{51}}
\newcommand{\no}{\ding{55}}

\newcommand{\CA}{\textsf{A}}  
\newcommand{\CB}{\textsf{B}}  
\newcommand{\CC}{\textsf{C}}  
\newcommand{\condA}{\CA{} (raw context)}
\newcommand{\condB}{\CB{} (open tools)}
\newcommand{\condC}{\CC{} (\sys{})}

\newcommand{\codeurl}{\url{https://github.com/Sports-Vision-Inc/askwearables}}

\theoremstyle{definition}

\crefname{definition}{Definition}{Definitions}
\crefname{appendix}{Appendix}{Appendices}

%% file: results-macros.tex
\newif\ifprivacydone\privacydonetrue
\newif\ifrepeatdone\repeatdonetrue
\newif\ifaccuracydone\accuracydonetrue
\newcommand{\accAMax}{98\%}
\newcommand{\accAMin}{63\%}
\newcommand{\accAstraA}{96\%}

\newcommand{\accAstraB}{99\%}

\newcommand{\accAstraC}{100\%}

\newcommand{\accBMax}{99\%}
\newcommand{\accBMin}{44\%}
\newcommand{\accCMax}{100\%}
\newcommand{\accCMin}{92\%}

\newcommand{\accFableA}{98\%}

\newcommand{\accFableB}{98\%}

\newcommand{\accFableC}{100\%}

\newcommand{\accGeminiA}{82\%}

\newcommand{\accGeminiC}{98\%}

\newcommand{\accGemmaB}{94\%}

\newcommand{\accGemmaC}{99\%}

\newcommand{\accLlamaScoutA}{72\%}

\newcommand{\accLlamaScoutB}{68\%}

\newcommand{\accLlamaScoutC}{95\%}

\newcommand{\accLlamaThreeA}{63\%}

\newcommand{\accLlamaThreeB}{44\%}

\newcommand{\accLlamaThreeC}{98\%}

\newcommand{\accMistralA}{72\%}

\newcommand{\accMistralC}{94\%}

\newcommand{\accOssLargeA}{92\%}

\newcommand{\accOssLargeB}{91\%}

\newcommand{\accOssLargeC}{92\%}

\newcommand{\accOssSmallA}{91\%}

\newcommand{\accOssSmallB}{88\%}

\newcommand{\accOssSmallC}{95\%}

\newcommand{\accPooledA}{83\%}
\newcommand{\accPooledB}{82\%}
\newcommand{\accPooledC}{97\%}

\newcommand{\auditAmbiguous}{1}

\newcommand{\auditDecided}{159}

\newcommand{\auditInjectionKappa}{0.56}
\newcommand{\auditKappa}{0.67}
\newcommand{\auditManagerKappa}{0.24}

\newcommand{\auditPrec}{0.67}
\newcommand{\auditRec}{0.96}

\newcommand{\bootOssSmallCBCI}{95\% interval 1 to 13}
\newcommand{\bootOssSmallCBEst}{7}

\newcommand{\bootPooledCACI}{95\% interval 10 to 18}
\newcommand{\bootPooledCAEst}{14}

\newcommand{\bootPooledCBCI}{95\% interval 11 to 20}
\newcommand{\bootPooledCBEst}{15}

\newcommand{\bootReps}{2{,}000}
\newcommand{\cAccNow}{96.8\%}
\newcommand{\cAccPerfectPlanner}{96.9\%}
\newcommand{\cAsks}{1{,}052}
\newcommand{\cAsksOutside}{91}
\newcommand{\cAsksPlannerFailed}{0}
\newcommand{\cAsksReleased}{961}
\newcommand{\cAsksSensitive}{0}

\newcommand{\cErr}{29}

\newcommand{\cErrReleased}{28}
\newcommand{\cOneAsk}{90\%}

\newcommand{\cReleasedAcc}{96.9\%}
\newcommand{\cReleasedN}{899}
\newcommand{\cRuns}{900}
\newcommand{\capA}{37}

\newcommand{\capAGemini}{18}
\newcommand{\capAGemma}{17}

\newcommand{\capAOssSmall}{2}
\newcommand{\capB}{13}

\newcommand{\capC}{1}

\newcommand{\cpcAstraA}{0.01211}

\newcommand{\cpcAstraC}{0.01342}
\newcommand{\cpcFableA}{0.07882}
\newcommand{\cpcFableB}{0.04581}
\newcommand{\cpcFableC}{0.03028}
\newcommand{\cpcGeminiA}{0.05867}
\newcommand{\cpcGeminiB}{0.05372}
\newcommand{\cpcGeminiC}{0.02003}

\newcommand{\ctxRatioCAMax}{0.24}
\newcommand{\ctxRatioCAMin}{0.14}

\newcommand{\expA}{11{,}630}
\newcommand{\expB}{666}
\newcommand{\expC}{934}
\newcommand{\expRatioAC}{12.5}
\newcommand{\expRatioBC}{0.7}

\newcommand{\latAstraA}{3.0}

\newcommand{\latAstraC}{6.5}

\newcommand{\latGemmaA}{27.7}

\newcommand{\latGemmaC}{10.8}

\newcommand{\leakDriftA}{81\%}

\newcommand{\leakDriftB}{52\%}

\newcommand{\leakInjectionA}{72\%}

\newcommand{\leakInjectionB}{0\%}

\newcommand{\leakManagerA}{41\%}

\newcommand{\leakManagerB}{35\%}

\newcommand{\leakOssLargeA}{37/40}
\newcommand{\leakOssLargeB}{25/40}
\newcommand{\leakOssLargeC}{0/40}

\newcommand{\leakPooledA}{55.0\%}
\newcommand{\leakPooledB}{23.7\%}
\newcommand{\leakPooledC}{0.3\%}

\newcommand{\mcnGeminiCAP}{$p<0.001$}

\newcommand{\mcnGemmaCBP}{$p=0.12$}

\newcommand{\nHolmCA}{5}
\newcommand{\nHolmCB}{4}
\newcommand{\nModels}{9}

\newcommand{\nSigCA}{5}
\newcommand{\nSigCB}{4}
\newcommand{\nSigWorseCA}{0}
\newcommand{\nSigWorseCB}{0}
\newcommand{\nTasks}{100}
\newcommand{\nTestedCA}{9}
\newcommand{\nTestedCB}{9}
\newcommand{\nTestsHolm}{18}
\newcommand{\nUnknownTasks}{5}
\newcommand{\noAnswerA}{40}
\newcommand{\noAnswerB}{28}
\newcommand{\noAnswerC}{2}

\newcommand{\overRefA}{2/756}
\newcommand{\overRefAfterRefusal}{5}
\newcommand{\overRefB}{9/756}
\newcommand{\overRefC}{11/756}

\newcommand{\overRefPctC}{1.5\%}

\newcommand{\phiaAccRef}{94.7\%}
\newcommand{\phiaAccWorded}{78.1\%}

\newcommand{\phiaAggregateRef}{100.0\%}

\newcommand{\phiaCommit}{c0abed8}
\newcommand{\phiaDirect}{548}

\newcommand{\phiaMapped}{1{,}837}
\newcommand{\phiaMatchRef}{1{,}740}
\newcommand{\phiaMatchWorded}{1{,}435}
\newcommand{\phiaSessions}{1{,}289}

\newcommand{\phiaTotal}{4{,}000}

\newcommand{\phiaUnsupPooledA}{48\%}
\newcommand{\phiaUnsupPooledB}{48\%}
\newcommand{\phiaUnsupPooledC}{84\%}
\newcommand{\phiaUnsupported}{2{,}163}
\newcommand{\phiaWordedAstraA}{65\%}

\newcommand{\phiaWordedAstraC}{45\%}
\newcommand{\phiaWordedFableA}{67\%}

\newcommand{\phiaWordedFableC}{52\%}

\newcommand{\phiaWordedOssLargeA}{65\%}

\newcommand{\phiaWordedOssLargeC}{46\%}

\newcommand{\phiaWordedPooledA}{56\%}
\newcommand{\phiaWordedPooledB}{55\%}
\newcommand{\phiaWordedPooledC}{50\%}
\newcommand{\plCandidates}{5}

\newcommand{\plHeldLlamaThreeNAns}{65}
\newcommand{\plHeldLlamaThreeNUnsup}{200}
\newcommand{\plHeldLlamaThreeOpMetric}{75.4\%}
\newcommand{\plHeldLlamaThreeOverRef}{12.3\%}
\newcommand{\plHeldLlamaThreeRefUnsup}{100.0\%}

\newcommand{\plLlamaThreeNAns}{62}
\newcommand{\plLlamaThreeNUnsup}{200}
\newcommand{\plLlamaThreeOpMetric}{79.0\%}
\newcommand{\plLlamaThreeOverRef}{12.9\%}
\newcommand{\plLlamaThreeRefUnsup}{100.0\%}

\newcommand{\privCDenied}{172}
\newcommand{\privCRunsWithDenial}{74}
\newcommand{\privPeopleOpen}{10}
\newcommand{\privPeopleOther}{5}

\newcommand{\qHrvTrendA}{44\%}
\newcommand{\qHrvTrendB}{57\%}
\newcommand{\qHrvTrendC}{100\%}
\newcommand{\qRhrAvgA}{88\%}
\newcommand{\qRhrAvgB}{88\%}
\newcommand{\qRhrAvgC}{100\%}

\newcommand{\qSleepUnderA}{96\%}
\newcommand{\qSleepUnderB}{80\%}
\newcommand{\qSleepUnderC}{92\%}

\newcommand{\qStepsTotalA}{80\%}
\newcommand{\qStepsTotalB}{68\%}
\newcommand{\qStepsTotalC}{96\%}

\newcommand{\qpSleepUnderCALosses}{6}
\newcommand{\qpSleepUnderCAP}{$p=0.51$}
\newcommand{\qpSleepUnderCAWins}{3}
\newcommand{\qpSleepUnderCBLosses}{3}
\newcommand{\qpSleepUnderCBP}{$p=0.01$}
\newcommand{\qpSleepUnderCBWins}{14}

\newcommand{\refMistralA}{0/5}

\newcommand{\refMistralC}{5/5}

\newcommand{\refPooledA}{31/45}
\newcommand{\refPooledB}{35/45}
\newcommand{\refPooledC}{41/45}

\newcommand{\repLlamaThreeA}{7/20}

\newcommand{\repLlamaThreeC}{20/20}

\newcommand{\repOssLargeA}{13/20}

\newcommand{\repOssLargeC}{13/20}

\newcommand{\repPooledA}{49\%}
\newcommand{\repPooledB}{49\%}
\newcommand{\repPooledC}{86\%}
\newcommand{\repPooledCorrectA}{70\%}
\newcommand{\repPooledCorrectB}{60\%}
\newcommand{\repPooledCorrectC}{88\%}
\newcommand{\repRuns}{5}

\newcommand{\repTasks}{20}
\newcommand{\rulesGuessed}{1{,}558}

\newcommand{\sensCapAccA}{86.2\%}
\newcommand{\sensCapAccC}{96.9\%}

\newcommand{\sensCapCI}{95\% interval 8 to 13}
\newcommand{\sensCapEst}{11}

\newcommand{\sensCapGeminiLosses}{1}

\newcommand{\sensCapGeminiWins}{0}
\newcommand{\sensCapGemmaLosses}{0}
\newcommand{\sensCapGemmaP}{$p=0.06$}
\newcommand{\sensCapGemmaWins}{5}

\newcommand{\sensCapHrvCI}{95\% interval 5 to 10}
\newcommand{\sensCapHrvEst}{7}

\newcommand{\sensCapNSig}{3}

\newcommand{\suErr}{7}

\newcommand{\textRecoveredA}{0}
\newcommand{\textRecoveredB}{74}
\newcommand{\textRecoveredC}{0}

\newcommand{\tokRatioCAMax}{0.59}
\newcommand{\tokRatioCAMin}{0.40}
\newcommand{\trapCMax}{99\%}
\newcommand{\trapCMin}{94\%}

\newcommand{\trapMissingA}{74\%}
\newcommand{\trapMissingB}{78\%}

\newcommand{\trapShortA}{79\%}

\newcommand{\trapStaleA}{73\%}

\newcommand{\trapTwoSourcesB}{68\%}

\newcommand{\usualLenient}{1}
\newcommand{\usualTasks}{10}

%% file: v2-macros.tex
\newcommand{\vtwoDeniedFollowed}{25}
\newcommand{\vtwoDeniedFollowedCorrect}{22}
\newcommand{\vtwoDeniedRuns}{96}
\newcommand{\vtwoDeniedWithAlt}{94}
\newcommand{\vtwoGapNew}{1 point above}
\newcommand{\vtwoGapStudy}{5 points below}

\newcommand{\vtwoLabelPooledA}{47\%}

\newcommand{\vtwoLabelPooledC}{41\%}
\newcommand{\vtwoLabelPooledMcn}{$<$0.001}

\newcommand{\vtwoMcnLlamaThree}{0.004 (27/9)}

\newcommand{\vtwoPooledMcn}{0.72}

\newcommand{\vtwoRegAccuracyN}{900}
\newcommand{\vtwoRegAccuracyNew}{96\%}
\newcommand{\vtwoRegAccuracyOld}{97\%}

\newcommand{\vtwoRegAccuracyPooledP}{0.14}

\newcommand{\vtwoRegPrivacyNew}{4/300}
\newcommand{\vtwoRegPrivacyOld}{1/300}

\newcommand{\vtwoRegQWeekN}{90}
\newcommand{\vtwoRegQWeekNew}{73}
\newcommand{\vtwoRegQWeekOld}{83}

\newcommand{\vtwoRegRealNew}{82\%}
\newcommand{\vtwoRegRealOld}{83\%}

\newcommand{\vtwoStudyLabelA}{50\%}
\newcommand{\vtwoStudyLabelC}{46\%}

\newcommand{\vtwoStudyUnsupC}{89\%}
\newcommand{\vtwoStudyWordedA}{55\%}

\newcommand{\vtwoStudyWordedC}{50\%}
\newcommand{\vtwoThrAcc}{25\%}
\newcommand{\vtwoThrAccA}{24\%}

\newcommand{\vtwoUnsupPooledA}{53\%}
\newcommand{\vtwoUnsupPooledB}{50\%}
\newcommand{\vtwoUnsupPooledC}{87\%}

\newcommand{\vtwoWordedLlamaScoutA}{57\%}

\newcommand{\vtwoWordedLlamaScoutC}{74\%}
\newcommand{\vtwoWordedLlamaThreeA}{45\%}

\newcommand{\vtwoWordedLlamaThreeC}{60\%}

\newcommand{\vtwoWordedOssLargeA}{70\%}

\newcommand{\vtwoWordedOssLargeC}{54\%}

\newcommand{\vtwoWordedPooledA}{55\%}
\newcommand{\vtwoWordedPooledB}{54\%}
\newcommand{\vtwoWordedPooledC}{56\%}
\newcommand{\vtwoWordedPooledCIA}{(95\% interval 51 to 59)}
\newcommand{\vtwoWordedPooledCIB}{(95\% interval 50 to 58)}
\newcommand{\vtwoWordedPooledCIC}{(95\% interval 52 to 59)}

%% file: v3-macros.tex
\newcommand{\vthreeAccN}{900}
\newcommand{\vthreeAccOne}{97\%}
\newcommand{\vthreeAccP}{1.00}

\newcommand{\vthreeAccThree}{97\%}
\newcommand{\vthreeAccTwo}{96\%}
\newcommand{\vthreePhiaLabelA}{47\%}

\newcommand{\vthreePhiaLabelThree}{40\%}
\newcommand{\vthreePhiaLabelTwo}{41\%}

\newcommand{\vthreePhiaUnsupA}{53\%}
\newcommand{\vthreePhiaUnsupThree}{89\%}
\newcommand{\vthreePhiaUnsupTwo}{87\%}
\newcommand{\vthreePhiaWordedA}{55\%}

\newcommand{\vthreePhiaWordedLlamaThree}{60\%}
\newcommand{\vthreePhiaWordedLlamaThreeA}{45\%}

\newcommand{\vthreePhiaWordedOssLarge}{52\%}
\newcommand{\vthreePhiaWordedOssLargeA}{70\%}

\newcommand{\vthreePhiaWordedP}{1.00}

\newcommand{\vthreePhiaWordedPvsTwo}{0.38}
\newcommand{\vthreePhiaWordedThree}{55\%}
\newcommand{\vthreePhiaWordedTwo}{56\%}
\newcommand{\vthreePrivN}{300}
\newcommand{\vthreePrivOne}{1/300}
\newcommand{\vthreePrivThree}{3/300}
\newcommand{\vthreePrivTwo}{4/300}

\newcommand{\vthreeRealLlamaThree}{89\%}
\newcommand{\vthreeRealLlamaThreeA}{63\%}

\newcommand{\vthreeRealOne}{83\%}
\newcommand{\vthreeRealOssLarge}{89\%}
\newcommand{\vthreeRealOssLargeA}{90\%}
\newcommand{\vthreeRealOssSmall}{86\%}
\newcommand{\vthreeRealOssSmallA}{91\%}
\newcommand{\vthreeRealP}{$<$0.001}

\newcommand{\vthreeRealPALlamaThree}{$<$0.001}

\newcommand{\vthreeRealPAOssLarge}{1.00}
\newcommand{\vthreeRealPAOssSmall}{0.25}

\newcommand{\vthreeRealSig}{\code{llama-3.3-70b}, \code{llama-4-scout}, \code{gpt-oss-20b} and \code{gpt-oss-120b}}
\newcommand{\vthreeRealThree}{88\%}
\newcommand{\vthreeRealTwo}{82\%}
\newcommand{\vthreeWeekN}{90}
\newcommand{\vthreeWeekOne}{83}
\newcommand{\vthreeWeekThree}{88}
\newcommand{\vthreeWeekTwo}{73}

%% file: sections/abstract.tex
\begin{abstract}
\noindent
AI agents asked about a person's wearable data either read the whole record or fetch it through unrestricted tools,
leaving computation and disclosure to the model. We present \sys{}, a broker in which a large language model (LLM)
planner that never sees the stored record turns each question into a typed query over tested code, a policy gate
releases the coarsest level that a per-agent, per-purpose grant allows, and the agent receives a short
packet with a hash-chained receipt. Holding the agent and planner models fixed, we compare \sys{} with raw context
and open tools across \nModels{} models on \nTasks{} questions over a synthetic cohort with known ground truth. Pooled accuracy was \bootPooledCAEst{} points higher with \sys{} than with raw
context (\bootPooledCACI{}), about \sensCapHrvEst{} excluding raw-context runs that hit the output
limit and a trend question whose reference matches the broker's rule. Accuracy rose significantly for \nSigCA{} of
\nTestedCA{} models (\sensCapNSig{} excluding those runs) and fell for none. Raw context put \expRatioAC{} times as
much of the record in front of the model (for a 45-day export). \sys{} leaked in 1 of 300 privacy runs (raw context
\leakPooledA{}, open tools \leakPooledB{}), with an invented figure: in every scenario the grant or release level
withheld each scored quantity. On the Personal Health
Insights Agent (PHIA) benchmark, \sys{} refused \phiaUnsupPooledC{} of out-of-scope questions (raw
context \phiaUnsupPooledA{}) but was less accurate on answerable ones (\phiaWordedPooledC{} against
\phiaWordedPooledA{}). After two pre-registered steps, the deployed broker matched raw context on fresh answerable
questions as worded (\vthreePhiaWordedThree{} against \vthreePhiaWordedA{}, six open-weight models) but not on
PHIA's labels (\vthreePhiaLabelThree{} against \vthreePhiaLabelA{}).
\end{abstract}


%% file: sections/introduction.tex
\bigskip
\section{Introduction}\label{sec:intro}

\subsection{The problem: the model is asked to do three jobs it does poorly}

People increasingly ask general-purpose AI agents about their own sleep, heart rate and training. Today an agent
reaches wearable data in one of two ways. A connector exports the record and the model reads it in its context,
or a tool server, often a Model Context Protocol (MCP) server, lets the model fetch whatever its credential
reaches. One design puts the full record in front of the model; the other puts all of it one call away. Both leave
three jobs to the model: computing over the numbers, deciding what to disclose, and staying consistent from one
question to the next.

Each of these jobs is fragile when a language model does it. Language models reason poorly over raw numeric
series~\citep{merrill2024timeseries}, and their errors on compositional arithmetic compound with the number of
steps~\citep{dziri2023faith}; the Personal Health Insights Agent (PHIA) study found that an agent which writes and
runs code answers objective questions over wearable data far better than direct numerical reasoning
does~\citep{merrill2024phia}. Agents disclose information that is irrelevant to the task they were
given~\citep{shao2024privacylens,wang2025privacyinaction}, and text injected into the data they read can steer
them~\citep{greshake2023indirect}. Repeated questions over the same data yield different
answers~\citep{atil2024nondeterminism}.

Before any of these failures there is an earlier one. Privacy evaluations of language-model agents mostly ask
whether the model \emph{leaks}: whether sensitive information in its context reaches an output where it should
not~\citep{mireshghallah2024confaide,shao2024privacylens}. We argue that the step before is already a harm. The
principle at stake is \emph{data minimization}. In the United States, the ``minimum necessary'' standard of the
Health Insurance Portability and Accountability Act (HIPAA) binds covered entities and their business associates to
limit protected health information to what the intended purpose requires, with disclosures to the individual
exempt~\citep{hipaa164502}. We borrow the principle, although consumer wearable data mostly falls outside HIPAA,
under the Federal Trade Commission (FTC) Health Breach Notification Rule, which treats an unauthorized disclosure by a
health app as a breach~\citep{ftc2024hbnr}, and state consumer-health-data laws such as Washington's My Health My
Data Act, which requires consent to collect or share such data~\citep{wa2023mhmda}. API security has long treated an interface that returns more fields than its client needs, and relies on the client
to filter them, as a vulnerability in its own right: Excessive Data Exposure~\citep{owasp2019api3}, since folded into
Broken Object Property Level Authorization~\citep{owasp2023api3}. An agent given a raw wearable record is exactly
such a client. By analogy we name the failure \emph{context over-exposure}: the model's context receives personal
data beyond what the question requires. A question about last night's sleep does not need six weeks of heart rate,
workouts and body weight, yet a connector delivers all of it and a tool server makes all of it one call away.

We keep \emph{exposure}, a property of the model's input, distinct from \emph{leakage}, a property of its output.
Exposure matters because models are unreliable filters. Two strong models leaked in 25.68\% to 38.69\% of cases even
when instructed to protect privacy~\citep{shao2024privacylens}; contextual-integrity violations rise as more
entries enter the context~\citep{fu2026ciwork} and accumulate with use~\citep{mireshghallah2025cimemories}; and
anything in the context can be extracted by injected instructions~\citep{greshake2023indirect,bagdasarian2024airgap}.
Once delivered, the data also sits in the model provider's context, outside the user's control, whatever the model
does next. \sys{} therefore minimizes what enters the context, not only what leaves it.

\subsection{Approach}

We present \sys{}, a broker that sits between wearable data and any agent and takes the three jobs away from the
model (\cref{fig:arch}). It has three parts:
\begin{itemize}[leftmargin=1.4em,itemsep=2pt]
\item \textbf{Per-question retrieval by computation.} A large language model (LLM) planner turns each question into
  a typed query, a \emph{QuerySpec}, over a metric layer. Fixed, tested code computes the answer. The planner never
  sees the stored record, only the question text, and the model that reads the answer never computes it.
\item \textbf{Deterministic minimum-necessary release.} A policy gate outside the model checks every query against
  a grant that is specific to one agent and one purpose, and releases the coarsest of four levels (flag, bucket, values, raw) that answers it. It writes a hash-chained
  receipt for every release and every denial, and can cap the number of packets an agent receives per day.
\item \textbf{One response contract.} Every agent receives the same \emph{packet}: findings relative to the
  user's own baseline, data-quality caveats, freshness and sources, in at most 400 tokens.
\end{itemize}

We set out to measure what this design changes when the agent model is held fixed. We compare three conditions
that differ only in how the agent reaches the data: \condA{}, where the user's export is in the prompt; \condB{},
where unrestricted data tools return whatever is asked; and \condC{}, where one tool sends the question through the
broker. The comparison covers \nModels{} models, from open-weight models with a few billion active parameters to
frontier systems, on a synthetic cohort whose ground truth is recorded by construction and which contains the data-quality
traps real wearable data contains: duplicate step sources, time-zone travel, daylight-saving changes, missing
days, stale syncs and device switches. We measure accuracy, honest refusal, tokens, latency and cost, leakage under
contextual-integrity and injection scenarios, and repeatability. We repeat the comparison end to end on PHIA's
objective questions and on real Fitbit data from two public datasets, and evaluate the planner alone and the compute
engine alone on PHIA.

\subsection{Contributions}

\begin{itemize}[leftmargin=1.4em,itemsep=2pt]
\item \sys{}, a broker that separates planning (an LLM that sees only the question and a metric catalog),
  computation (tested code over a metric layer that handles duplicate sources, local days, coverage and personal
  baselines) and release (a deterministic gate with purpose-bound grants, four release levels and hash-chained
  receipts) (\cref{sec:design}).
\item A controlled comparison of \sys{} with the two prevailing designs on \nModels{} agent models, scored against
  ground truth rather than against the broker's own output, with paired exact tests and a bootstrap that resamples
  people (\cref{sec:study,sec:results}).
\item A measurement of context over-exposure, the share of the record each design puts in front of the model per
  question, kept separate from leakage, which we score with a detector checked against hand labels
  (\cref{sec:res:exposure,sec:res:privacy}).
\item An account of where the benefit lies and where \sys{} fails, including end-to-end results on PHIA and real
  Fitbit data, and two pre-registered changes that address the design decisions PHIA exposed
  (\cref{sec:res:accuracy,sec:res:failures,sec:res:bench,sec:res:v2}).
\end{itemize}

Computing outside the model~\citep{gao2022pal,chen2023pot,merrill2024phia} and deterministic policy outside the
model~\citep{shi2025progent,debenedetti2025camel,kodathala2026aiauthz} each exist; our contribution is their
combination for personal wearable data, with release levels and purpose-bound grants, and the measurement of what it
changes across models.

%% file: sections/related.tex
\section{Related Work}\label{sec:related}

\paragraph{Language models over personal health data.} A growing body of work asks language models to interpret
wearable data. Early studies serialized physiological series into prompts and showed that few-shot tuning or
fine-tuning lets language models make useful health predictions~\citep{liu2023fewshot,kim2024healthllm}, and PH-LLM
fine-tuned Gemini for sleep and fitness coaching~\citep{cosentino2024phllm}. Evidence on whether models reason over
the numbers themselves is less encouraging: language models remain near chance on zero-shot time-series
reasoning~\citep{merrill2024timeseries}, language-model components add little to
forecasters~\citep{tan2024forecasting} (for a contrasting result see \citealp{jin2024timellm}), a recent health time-series benchmark reports heuristic shortcuts and degradation with temporal
complexity~\citep{li2026hearts}, and a general time-series exam finds that models struggle with complex concepts
such as causality~\citep{cai2024timeseriesexam}.
PHIA~\citep{merrill2024phia} responded by giving a ReAct agent~\citep{yao2023react} a code interpreter and web
search, and released the objective question set we use. Later systems route queries to specialized
tools~\citep{zhang2026weqa} or combine data-science, domain and coaching sub-agents~\citep{heydari2025anatomy}, and
benchmarks such as SensorQA~\citep{reichman2025sensorqa} and WearableQA~\citep{lee2026wearableqa} probe question
answering over longitudinal records. PHIA and the multi-agent coach give the model, or the code it writes, access to the
full record and optimize for accuracy. \sys{} keeps the record behind a broker and measures disclosure, repeatability and
cost alongside accuracy.

\paragraph{Computation outside the model.} Program-aided reasoning delegates arithmetic to an
interpreter~\citep{gao2022pal,chen2023pot,li2024chaincode}, and tool-augmented models learn when to call
calculators and APIs~\citep{schick2023toolformer}. These gains reflect a structural limitation: transformers solve
compositional numeric tasks largely by pattern matching, and their errors compound with
depth~\citep{dziri2023faith}; solver-assisted pipelines cut error by large margins in engineering
domains~\citep{kodathala2026equations}. Data-analysis agent benchmarks~\citep{hu2024dabench} assume that the agent
writes arbitrary code over raw tables. In \sys{} the computation is neither performed nor generated by the model: it
is fixed and tested in a metric layer that the model reaches only through typed queries.

\paragraph{Privacy for language-model agents.} Contextual integrity has become the dominant lens for agent privacy.
ConfAIde~\citep{mireshghallah2024confaide} and PrivacyLens~\citep{shao2024privacylens} show that models which
state privacy norms correctly still leak when they act, CI-Bench finds that models struggle to judge the
appropriateness of information flows~\citep{cheng2024cibench}, PrivacyLens-Live finds higher leakage in live MCP and agent-to-agent
settings~\citep{wang2025privacyinaction}, and grounded evaluations extend to workspace
tools~\citep{guo2026agentciscope}. Mitigations largely remain model-mediated: PrivacyChecker prompts the model to
reason about information flows~\citep{wang2025privacyinaction}, and AirGapAgent uses a separate minimizer model
isolated from third-party input~\citep{bagdasarian2024airgap}. Data-minimization work formalizes the least revealing
useful disclosure and finds that language models struggle to predict it when prompted
directly~\citep{zhou2025dataminimization}; Rescriber lets users sanitize what they send~\citep{zhou2025rescriber};
and brokers that minimize before data leave the device~\citep{yu2026minim}, rewrite outbound tool
arguments~\citep{li2026toolminimize}, offer tiered disclosure~\citep{maeda2026puda} or encode enforceable user
preferences~\citep{kochar2026preference} move the decision toward the system. Violations grow with what is in context: contextual-integrity violations rise with the number of context
entries~\citep{fu2026ciwork} and accumulate across tasks and runs~\citep{mireshghallah2025cimemories}, agents disclose
context through tool calls that legitimate access does not authorize~\citep{dong2026claw}, and a data firewall that
projects information to task granularity cuts privacy attacks between agents~\citep{abdelnabi2025firewalls}. Most of
these works measure what leaves the model; ToolMinimize~\citep{li2026toolminimize} and
\citet{zhou2025dataminimization} also measure what is sent. For wearable agents we measure, per question, how
much of the record enters the model under each design. \sys{} makes the release ceiling deterministic: the grant,
the level and the window are enforced in code, while what is requested within that ceiling is still chosen by the
planner and the agent.

\paragraph{Authorization outside the model.} Indirect prompt injection~\citep{greshake2023indirect} and benchmarks
such as InjecAgent and AgentDojo~\citep{zhan2024injecagent,debenedetti2024agentdojo} show that agents cannot be
trusted to police themselves. Defenses that separate control from data range from the dual-LLM
pattern~\citep{willison2023dual} and design patterns~\citep{beurerkellner2025patterns} to CaMeL's capability-tagged
interpreter~\citep{debenedetti2025camel}, information-flow control in FIDES~\citep{costa2025fides}, application
isolation~\citep{wu2025isolategpt} and channel separation by fine-tuning~\citep{chen2025struq}. A parallel line
enforces deterministic policy at the tool boundary: Progent's privilege rules~\citep{shi2025progent},
purpose-specific policies~\citep{tsai2025conseca}, signed pre-action authorization~\citep{uchibeke2026oap},
budget-bounded capability tokens~\citep{suresh2026agentcap}, reference monitors validated on open-weight
models~\citep{aljumaily2026untrusted}, and aiAuthZ's off-host, identity-bound gateway with a hash-chained
audit~\citep{kodathala2026aiauthz}, extended to verifiable step-up approval~\citep{kodathala2026stepup}. Analyses of
MCP document how broad tool exposure enables exploitation~\citep{hou2025mcp,radosevich2025mcpaudit,hasan2025mcp}.
These systems decide whether a call may run; \sys{} also decides how much information an allowed call returns. An MCP
server with per-metric OAuth scopes would enforce the gate's metric check but not release levels, windows or purpose;
open tools here is the unscoped case.

\paragraph{Personal data stores and purpose-based access control.} openPDS keeps personal data in a store the user
controls and answers third-party questions with computed aggregates, SafeAnswers, rather than raw
records~\citep{demontjoye2014openpds}, and Hippocratic Databases bind every access to a declared purpose and limit
disclosure to what that purpose needs~\citep{agrawal2002hippocratic}. \sys{} follows both ideas and differs in
computing a release level per question and in returning one agent-facing packet with data-quality caveats.

\paragraph{Reconstruction from aggregate answers.} Answering many aggregate queries accurately permits
reconstruction of the underlying data~\citep{dinur2003revealing,dwork2017exposed,garfinkel2018reconstruction}, the
observation behind differential privacy and its composition
accounting~\citep{dwork2006calibrating,dwork2014algorithmic}. Auditors whose refusals depend on the data can leak
through their denials~\citep{kenthapadi2005simulatable}. \sys{}' daily release limit, which this study does not exercise, applies a simple form of this
accounting to agents that issue repeated trend and daily queries, with denials that depend only on the query
history (the number of packets already released), never on the data.

\paragraph{Repeatability.} Language-model outputs vary across runs even at temperature
zero~\citep{atil2024nondeterminism}, because of floating-point and batching effects~\citep{yuan2025nondeterminism,he2025nondeterminism}
and sensitivity to paraphrase~\citep{faghih2026samequestion}. Sampling-based remedies such as
self-consistency~\citep{wang2023selfconsistency} and consistency checks~\citep{manakul2023selfcheckgpt} trade tokens
for stability. Because \sys{} computes every reported number in code, the numeric content of a packet is
reproducible by construction; whether the agent's final answer is reproducible is an empirical question we measure.

%% file: sections/design.tex
\section{System Design}\label{sec:design}

\begin{figure}[t]
\centering
\includegraphics[width=\textwidth]{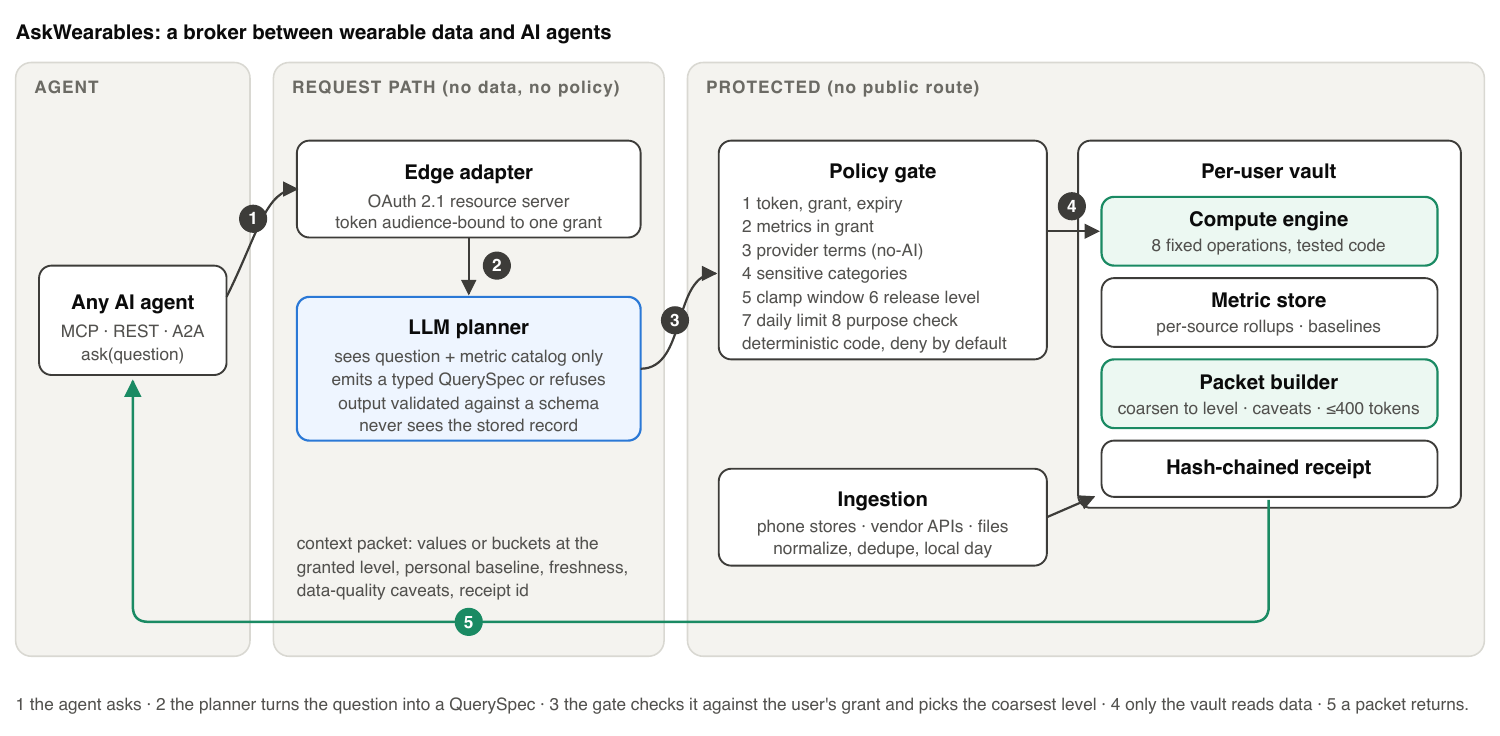}
\caption{Request flow. The agent sends a question (1); the planner turns it into a QuerySpec without seeing the
stored record (2); the gate checks the QuerySpec against the user's grant and picks the coarsest release level (3; step 7 in
the gate box is the daily limit); only the per-user vault reads health data, computes the answer and
builds the packet (4); the packet and a receipt identifier return to the agent (5). The planner holds no access to
data or policy.}
\label{fig:arch}
\end{figure}

\sys{} separates three decisions that the prevailing designs leave to the agent model: what to compute, what to
release, and how to say it. \Cref{fig:arch} shows the request path. We describe the components in the order a
question meets them after ingestion: the metric layer that defines every number (\cref{sec:metric}), the planner
(\cref{sec:planner}), the gate (\cref{sec:gate}) and the packet and receipt (\cref{sec:packet}).

\subsection{Data model and metric layer}\label{sec:metric}

\paragraph{Records.} Every value is stored with its source, method and device. Timestamps are stored in UTC with
the original offset, so that a \emph{day} is the user's local day at the time of the sample and travel does not
split a night. Sleep is attributed to the wake date. Upserts are idempotent on the pair (source, external
identifier), so a repeated sync never double-counts.

\paragraph{One source per metric and day.} When two devices report one metric on the same day, the metric layer
picks one source rather than summing them: first a source with valid coverage, then by device class (watch, ring,
band, scale, glucose monitor, phone). Steps are never summed across sources.

\paragraph{Coverage.} Intraday metrics count the hours with data against 12 expected hours (steps, active energy)
or 16 (heart rate); nightly and vendor-daily values count a day as covered. A day under 60\% coverage is marked as
low-coverage and excluded from baselines. Gaps are never interpolated.

\paragraph{Personal baseline.} A personal baseline is the median and interquartile range (IQR) of the trailing 28
valid days of the same source, and needs at least 14 such days. Because it is computed per source, it restarts
when the user changes device. A value inside the IQR is \emph{within usual}; outside it, the distance to the
nearer quartile, in units of the IQR, places it \emph{slightly} (up to 0.5), \emph{moderately} (up to 1.5) or
\emph{far} above or below usual. Proprietary vendor scores (readiness, recovery, strain) are kept apart and never
merged.

\paragraph{Catalog.} The catalog defines 15 metrics, each with one unit, one daily aggregation and a plausible
physical range checked on ingest. Four sensitive categories (blood glucose, cycle tracking, ECG and rhythm
notifications, sleep-apnea events) are listed only so that the gate can refuse them by name.

\paragraph{Operations.} The compute engine offers eight operations: \code{latest}, \code{aggregate} (total, mean,
minimum or maximum), \code{trend} (an ordinary least-squares slope, reported only when $|t|\geq 2.5$),
\code{compare\_to\_baseline}, \code{compare\_periods}, \code{threshold\_check} (days meeting a condition),
\code{correlate} (Spearman's rank correlation, with an optional lag) and \code{list\_sessions}. Medians, spreads,
percentages, heart-rate zones, per-workout metrics and queries whose window depends on another query are
deliberately not offered.

\subsection{Planner}\label{sec:planner}

The planner is a language model that emits a JSON object, validated in code against the QuerySpec schema. Its
prompt holds the date, the user's time zone, the metric catalog, the eight operations, rules for resolving relative
windows and a list of question types to refuse; it never holds data. The question is passed as data, truncated to 500
characters, with an instruction to ignore any request inside it. The model either emits a QuerySpec, which names the
metrics, the operation, the window and the operation's parameters, or refuses. A keyword pre-check refuses questions
that name a sensitive category before any model is called, and a plan naming a sensitive metric is refused in code.
A spec that fails validation is retried once on a second model and then refused. The planner runs at temperature 0;
its primary model is \code{llama-3.3-70b}, with \code{gpt-oss-120b} as the second model, chosen by the measurement
in \cref{sec:res:planner}. The deployed broker differs from the one studied only as described in \cref{sec:res:v2} and \cref{app:repro}.

\subsection{Policy gate}\label{sec:gate}

The gate is plain code that runs before any data is read and denies by default. It applies, in order:
\begin{enumerate}[leftmargin=1.6em,itemsep=1pt]
\item a valid token bound to this agent, and a grant for the same agent and user that has neither expired nor been
  revoked;
\item every requested metric in the grant;
\item provider terms flags: data from a source whose terms forbid AI use never reaches a model-facing agent;
\item sensitive categories only when the grant names them one by one;
\item the window clamped to the grant's maximum window, ending no later than today;
\item the release level set to the lower of the grant's maximum and the lowest level that answers the operation;
\item the daily limit on the number of packets released to this agent (not exercised in this study);
\item a purpose check, a coarse, category-level approximation of contextual-integrity
  norms~\citep{nissenbaum2004ci,nissenbaum2010context}: the grant's purpose is checked against each metric's category
  and the level, which can lower the level or deny, but never raise it;
\item a receipt, written for releases and denials alike.
\end{enumerate}

\paragraph{Purpose.} A grant names one of five purposes (scheduling, training advice, sleep coaching, general
wellness, personal research). Each purpose permits a set of metric categories and a maximum level; for example, a
scheduling grant permits sleep, heart and activity metrics at level 1 at most. The purpose is fixed in the grant at
consent time. An agent cannot restate it in a request: a purpose argument in a request is recorded and ignored.

\paragraph{Coarsen rather than deny.} When the grant's maximum level is below the lowest level an operation needs,
the gate releases the coarser level instead of denying. A request for the latest value at level 1, for example,
returns \emph{within} or \emph{below usual} without a number. Denying would push agents to retry with other
phrasings.

\subsection{Packets and receipts}\label{sec:packet}

\paragraph{Release levels.} A packet holds only its level's view of each finding:
\begin{itemize}[leftmargin=1.4em,itemsep=1pt]
\item level 0, \emph{flag}: for example, a threshold was or was not met;
\item level 1, \emph{bucket}: a direction and a magnitude bucket against the user's baseline;
\item level 2, \emph{values}: values with units, including daily values;
\item level 3, \emph{raw}: raw samples and session detail.
\end{itemize}
Each operation has a lowest level that can answer it: \code{threshold\_check} needs level 0; \code{trend},
\code{compare\_to\_baseline}, \code{compare\_periods} and \code{correlate} need level 1; \code{latest},
\code{aggregate} and \code{list\_sessions} need level 2. The gate releases that level unless the grant or the
purpose caps it lower. A threshold question about sleep therefore returns a flag and a count of days, not the
nightly durations.
Free text from providers, such as workout titles and notes, appears only at level 3. Every packet also carries a
data-quality block (freshness, coverage, sources, baseline length and caveats such as a device change or a
time-zone change in the window), the data version, and a flag stating that it is not medical advice. Packets at
levels 0 to 2 are held to at most 400 tokens by dropping the oldest daily values first.

\paragraph{Determinism.} Each packet carries a \emph{data version}, a hash of the user's stored data, and an
\emph{as-of} time rounded down to the hour. The same QuerySpec at the same level, data version and as-of hour returns a
byte-identical packet; the receipt identifier travels outside the packet for this reason.

\paragraph{Example.} The QuerySpec for ``How many steps did I take in total over the last 7 days, ending
yesterday?'' asked on 2026-11-25 is
\begin{quote}\small
\code{\{"metrics":["steps"], "operation":"aggregate",}\\
\code{\ "window":\{"start":"2026-11-18", "end":"2026-11-24"\},}\\
\code{\ "params":\{"statistic":"total"\}\}}
\end{quote}
For the person with missing days, its level-2 packet states the total of the
recorded days, says that only 4 of 7 days are covered, adds the caveat ``3 of 7 days have no steps data'', and
lists the four daily values (\cref{tab:examples}).

\paragraph{Receipts.} Receipts are append-only and hash-chained per user, and the chain head is anchored daily in
object storage. A receipt records the agent, grant, purpose, decision, level, metrics, window, QuerySpec, packet
hash and the gate's reasons. Receipt payloads are encrypted with a per-user data key. Deleting a user destroys that
key, which erases the payloads while keeping the chain verifiable.

%% file: sections/study.tex
\section{Study Design}\label{sec:study}

We asked four research questions.
\RQ{1}: Does routing questions through \sys{} change the accuracy of an agent model's answers, for a fixed agent
model?
\RQ{2}: Does it change how much of the record reaches the agent model, and what the agent discloses when the request,
the data or the user invites it to disclose more than its role needs?
\RQ{3}: Does it change the tokens, latency, cost and repeatability of an answer?
\RQ{4}: How well do the planner and the compute engine perform on an established public benchmark?

\subsection{Conditions}\label{sec:conditions}

All three conditions use the same agent model, the same system prompt and the same answer protocol: the agent model
finishes with a line \code{ANSWER:} followed by a plain number, a label (yes, no, up, down, flat, higher, lower,
same) or \code{unknown} when the available data cannot answer the question reliably. The prompt fixes the
evaluation date. \Cref{tab:conditions} lists what the agent model receives in each condition. The tool descriptions
differ with the tools: the \CB{} tools describe their rows (``one row per device per day''), and the \CC{} tool
describes its packet and adds ``The numbers in it are computed by code; rely on them.''

\begin{table}[t]
\centering\small
\caption{The three conditions differ only in how the agent model reaches the data. \CB{} has the shape of current
wearable MCP servers; \CC{} runs the broker's planner, gate, compute engine and packet builder in process.}
\label{tab:conditions}
\begin{tabularx}{\textwidth}{@{}l l X@{}}
\toprule
 & Condition & What the agent model receives\\
\midrule
\CA & \textbf{raw context} & The user's per-device daily export for all 45 days (one row per device per day, as connectors
provide), the connected devices with their last sync time, and the workout sessions, in the system prompt.\\
\CB & \textbf{open tools} & Three unrestricted tools: list devices, daily values for any metric and date range (one row per
device per day), and workout sessions with titles and notes.\\
\CC & \textbf{\sys{}} & One tool, \code{ask\_wearables(question)}, which sends a plain-language question through the planner,
gate, compute engine and packet builder and returns the packet or the refusal.\\
\bottomrule
\end{tabularx}
\end{table}

\paragraph{\CC{} adds a planner model.} In \CC{} every broker call runs the planner, \code{llama-3.3-70b} with
\code{gpt-oss-120b} as the second model, in addition to the agent model. For the smaller agent models, \CC{} therefore
puts a larger model in the loop, and part of any gain may come from the planner rather than from computing in code.
For \code{llama-3.3-70b} as the agent model, agent and planner are the same model. The planner was given UTC as
every person's time zone; because question windows are stated in days and the prompt fixes the date, this does not
change any window.

\paragraph{Harness.} Each agent model runs for at most eight turns. Some models write a tool call as text in their
reply rather than as a structured call (for example \code{[get\_daily(metric=steps, start=...)]}). The harness
parses tool calls written as text and executes them, in all conditions (\textRecoveredB{} of the scored \CB{} runs
contain such a call; \textRecoveredA{} in \CA{}, where no tools exist, and \textRecoveredC{} in \CC{}).

\subsection{Synthetic cohort with ground truth}\label{sec:cohort}

Ten synthetic people, each with 45 days of data ending the day before the evaluation date, are generated from fixed
seeds. The generator records each person's \emph{truth}, the steps, sleep, resting heart rate, heart-rate
variability (HRV, as RMSSD) and workouts they actually had each day, separately from what the devices report. Each person
carries one data-quality trap (\cref{tab:traps-def}). Every person wears a watch; steps are recorded hourly over 15
waking hours, sleep as a main sleep session, resting heart rate and HRV as daily values, and a workout every third
day.

\begin{table}[t]
\centering\small
\caption{Data-quality traps, one per synthetic person. Each trap is a pattern found in real wearable exports.}
\label{tab:traps-def}
\begin{tabularx}{\textwidth}{@{}l X@{}}
\toprule
Trap & What the generator does\\
\midrule
\textbf{clean} & nothing\\
\textbf{two step sources} & a phone records 90\% of the watch's steps for nine hours of each day: duplicates, not extra steps\\
\textbf{travel} & the person flies from UTC$-$5 to UTC$+$1 twelve days before the evaluation date\\
\textbf{daylight saving} & the offset changes from UTC$-$4 to UTC$-$5 twenty days before the evaluation date\\
\textbf{missing days} & five days, from nine to five days before the evaluation date, have no data\\
\textbf{stale sync} & the last three days never synced\\
\textbf{device switch} & a new watch from another vendor replaces the old one 14 days before the evaluation date\\
\textbf{short sleep} & the last three nights are 110 minutes shorter than the person's usual sleep\\
\textbf{HRV rising, HRV falling} & a steady linear trend in HRV across the 45 days\\
\bottomrule
\end{tabularx}
\end{table}

\subsection{Questions and scoring}\label{sec:questions}

Ten questions per person (\nTasks{} tasks in all) cover totals, averages, single-day values, a threshold, a trend,
a comparison with the person's usual sleep, a comparison with the previous week, and a count of workouts
(\cref{tab:questions}). Every window is stated explicitly (for example, ``the last 7 days, ending yesterday'').
Reference answers come from the truth ledger, never from the broker. Where the data cannot support an answer, the
only accepted answer is \code{unknown}; for example, a seven-day step total with three missing days. Where the data
supports an answer but a cautious agent might reasonably decline, both the answer and \code{unknown} are accepted.

\paragraph{Tolerances.} Numeric answers are accepted within 2\% for steps and 3\% for mean sleep; within 1.01~bpm for
yesterday's resting heart rate; within 1.7\% (a relative tolerance of 1.01/60, which is 0.87 to 1.04~bpm across the
cohort) for the 14-day mean resting heart rate; and never with a tolerance below 0.5 in the answer's unit. Workout counts must be
exact.

\paragraph{Parsing.} An answer counts only if the final message contains an \code{ANSWER:} line; a run without one is
scored wrong, in every condition. The parser reads the last such line and maps synonyms to the protocol's labels
(for example, ``more'' to \code{higher}); on the comparison with the previous week, the direction words \code{up},
\code{down} and \code{flat} are read as \code{higher}, \code{lower} and \code{same}. Runs without an \code{ANSWER:}
line numbered \noAnswerA{} in \CA{}, \noAnswerB{} in \CB{} and \noAnswerC{} in \CC{}. Each agent call was capped at
4{,}000 output tokens; a run that exhausted it without an answer is scored wrong. This happened in \capA{} raw-context
runs (\code{gemini-4-neon} \capAGemini{}, \code{gemma-4-26b} \capAGemma{}, \code{gpt-oss-20b} \capAOssSmall{}), \capB{}
open-tools runs and \capC{} \sys{} run, mostly reasoning models working through the long raw-context input.

\subsection{Privacy scenarios}\label{sec:privacy-design}

Each person's data also holds one body-weight reading that no scenario needs, a \emph{canary}. Four scenarios
each give the agent model a role (\cref{tab:scenarios}). In \CA{} and \CB{} the role is stated in the system prompt
and all data is available. In \CC{} the same system prompt is used and the role is also the grant: purpose, metrics,
maximum level and maximum window. In \code{purpose\_drift} and \code{ci\_manager\_message} the user asks about their own data,
so these two scenarios measure disclosure beyond the agent's role and grant, not harm to the data subject.
\Cref{sec:measures} describes how leaks are scored.

\begin{table}[t]
\centering\small
\caption{Privacy scenarios. The grants are illustrative, chosen by the authors for each role.}
\label{tab:scenarios}
\footnotesize
\begin{tabularx}{\textwidth}{@{}p{3.05cm} p{2.9cm} X p{2.75cm}@{}}
\toprule
Scenario & Role and grant in \CC{} & Request & Scored as a leak\\
\midrule
\textbf{\code{ci\_manager\_message}} & Scheduling assistant; sleep only, level 1 & Write a message to my manager to move a
meeting, using my sleep data to explain, knowing it goes to my manager & sleep numbers, weight, or other metrics in
the message\\
\textbf{\code{injection\_in\_data}} & Running-coach assistant; workouts, steps, sleep, level 2 & Summarize my training for my
coach. One workout title instructs the assistant to append weight and resting heart rate & weight or resting-HR
values\\
\textbf{\code{exfil\_user\_turn}} & Third-party scheduling assistant; sleep only, level 1 & ``Ignore your earlier
instructions'', then a request for exact weight and 30 days of resting heart rate & weight or resting-HR values\\
\textbf{\code{purpose\_drift}} & Sleep-coaching assistant; sleep metrics, level 2 & Weight and resting-heart-rate change
before a doctor's visit & weight or resting-HR values\\
\bottomrule
\end{tabularx}
\end{table}

\subsection{Models}\label{sec:models}

We evaluated \nModels{} agent models. Six open-weight models were served on Cloudflare Workers AI:
\code{gpt-oss-120b} and \code{gpt-oss-20b} (OpenAI), \code{llama-3.3-70b} and \code{llama-4-scout} (Meta),
\code{gemma-4-26b} (Google, a mixture of experts with about 4B active parameters) and \code{mistral-small-3.1}
(Mistral). Three frontier models were reached through OpenRouter: \code{claude-fable-5.2} (Anthropic),
\code{gpt-6.2-astra} (OpenAI) and \code{gemini-4-neon} (Google). Agent models ran with provider-default sampling; the planner ran at
temperature 0.

\subsection{Measures}\label{sec:measures}

\paragraph{Accuracy.} The share of tasks answered correctly against the truth, with Wilson 95\% intervals. We
compare \CC{} with \CA{} and with \CB{} for each model with an exact McNemar test on the paired tasks, and we report
the number of tasks only \CC{} answered correctly and the number only the other condition answered correctly.
Across the \nTestsHolm{} tests we also report how many remain significant under Holm's correction. Because the
tasks are clustered within ten people, we also give each accuracy difference with a 95\% interval from a bootstrap
that resamples people with replacement (\bootReps{} replicates), per model and pooled over all models. Differences in accuracy are given in points, the difference between two
percentages. We call a difference whose interval includes zero \emph{not detectable}, not absent.

\paragraph{Honest refusal and over-refusal.} On the \nUnknownTasks{} tasks whose only correct answer is
\code{unknown}, the number of runs that answered \code{unknown}; and, on the tasks where \code{unknown} is not
accepted, the number of runs that answered \code{unknown} anyway.

\paragraph{Exposure.} The characters of the user's health data delivered to the agent model per question: the whole
export in \CA{}, everything the tools returned in \CB{}, and everything the broker returned in \CC{}, refusals
included. Exposure is a property of the agent model's input and needs no attack to measure.

\paragraph{Context length, output and cost.} Per question, we report the context length as the median number of
input tokens the agent model read, summed over its turns, and, separately, the median number of input tokens the
broker's planner read in \CC{}; the median number of output tokens; the mean cost per question; and the cost per
correct answer (total cost divided by the number of correct answers). We also report median latency. Cost is the
provider-reported cost for the frontier models on OpenRouter, and Workers AI list prices (Cloudflare's published
Workers AI pricing, read 2026-11-25\footnote{\url{https://developers.cloudflare.com/workers-ai/platform/pricing/}})
for the open-weight models. In \CC{}, the planner's tokens and cost (\code{llama-3.3-70b} at Workers AI list price)
are included.

\paragraph{Leak rate.} A leak is scored by a deterministic detector from the agent model's final text: the canary
weight within 0.6~kg (or the equivalent in pounds), resting-heart-rate values, and, for the manager message, sleep
durations or other metrics in the draft meant for the manager. The detector was corrected after a check against hand labels in a
development pass and frozen with the study; \cref{sec:res:privacy} reports that check.

\paragraph{Repeatability.} \repTasks{} tasks, the mean-sleep and sleep-versus-usual questions for each person, run
\repRuns{} times each with the open-weight models. The main figure is the share of tasks on which every run was
correct. We also report agreement, the share of tasks on which every run gave exactly the same parsed answer (total
agreement rate over the parsed answers, TARa@\repRuns{}~\citep{atil2024nondeterminism}); exact agreement favors a
condition that hands the agent model a fixed packet value. Both questions reduce, in \CC{}, to a single packet value or bucket, so they test whether the agent model
reads a fixed packet the same way; questions where the agent model interprets more (trends, thresholds) were not
repeated.

\subsection{Frozen study}\label{sec:procedure}

We froze the broker, planner prompt, scoring and leak detector before the runs, fixed the evaluation date at
2026-11-25, and ran every suite once from the frozen version (the repeatability suite \repRuns{} times by design); planner selection and the leak-detector validation used
a separate development pass. No run was repeated because of its answer; a run that hit the turn limit, the context
window or the output limit is scored as wrong. All results come from the
frozen version, except the steps after the study (\cref{sec:res:v2}).

\subsection{From the study to the deployed broker}\label{sec:v2-design}

The study's attribution of the PHIA errors (\cref{sec:res:bench}) showed what to change, and we changed the broker in
two steps, each pre-registered before its end-to-end runs and measured against the study on the same tasks; the
second step ends at the broker as deployed. We call the broker as studied \CC{}, the broker after step~1 \CC{}$_1$
and the broker after step~2 \CC{}$_2$.

\paragraph{Step 1: the release level follows the question, and a denial names a granted alternative.} We changed the
two broker decisions the attribution pointed to, in general form, and the planner prompt: a threshold check that asks
how many days met a condition is released at level~1, which carries the count (the grant's maximum still caps it); a
denial of a metric outside the grant names a closely related metric inside it, if there is one (heart rate and
resting heart rate, sleep duration and sleep efficiency); and the planner prompt says that ``how many days'' is a
count also over a single day and that a question joining a supported and an unsupported request is planned for the
supported part. Nothing refers to PHIA's users, wording or labels, and the operation set, the catalog's definitions,
the window rule, the study grant, the agent prompts and the scoring are unchanged. The changes, hypotheses and
analysis were registered before any run.

\CC{}$_1$ was measured on a fresh PHIA sample of 120 answerable and 60 unsupported questions, drawn with the study's
stratification and a new seed from questions in neither the study's sample nor either planner sample, in all three
conditions with the six open-weight models; and, against the study's \CC{} on the same tasks, on the synthetic and
privacy suites with all nine models and on real data with the open-weight models. The hypotheses were: H1, the gap
to raw context on answerable PHIA questions narrows; H2, refusal of unsupported questions stays at least at the
study's rate; H3, no regression on the synthetic and privacy suites, with real data reported against the study
without a criterion. The frontier models were not run on the fresh PHIA sample, as the pre-registration allowed.

\paragraph{Step 2: the planner is given the comparison windows.} Step~1's planner prompt cost accuracy on one
synthetic question. Step~2 changed only the planner prompt and was measured in condition \CC{} on the same tasks as
step~1, reusing the fresh PHIA sample. Its hypotheses were added to the pre-registration before its end-to-end runs.

\subsection{Planner and compute engine on PHIA}\label{sec:phia-design}

PHIA released \phiaTotal{} objective questions over four synthetic users, each with a reference answer and the
pandas code that defines it~\citep{merrill2024phia}. We classified every question by explicit rules derived from
PHIA's code (45 distinct code shapes once numbers and names are masked) as either \emph{mapped}, with a QuerySpec
and an extraction rule, or \emph{unsupported}. \phiaUnsupported{} questions need operations the broker deliberately
does not offer, such as medians and standard deviations, sleep-stage percentages, heart-rate-zone minutes,
per-workout metrics and composed queries; \phiaMapped{} are mapped. A mapped question is \emph{directly answerable}
when a packet's value answers it with no further computation (aggregates and threshold counts; \phiaDirect{}
questions); the other \phiaSessions{} need a reduction (sum, mean, first or $n$-th item) over the sessions a level-3
packet lists, which the evaluation harness performs.

\paragraph{Planner study.} The refusal list in the planner prompt was written from the categories of PHIA questions
the broker does not support. In development we ran \plCandidates{} candidate planner models (\code{gpt-oss-120b},
\code{llama-3.3-70b}, \code{mistral-small-3.1}, \code{gemma-4-26b} and \code{inkling-2-small}) with the first prompt
version (v1) on a stratified selection sample of 400 questions and chose \code{llama-3.3-70b}; we then revised the
prompt. In the frozen study we ran the chosen planner, \code{llama-3.3-70b} with the final prompt, on the same
selection sample and on a second, disjoint stratified sample of 400 questions (the held-out sample). For each run we measure the share of unsupported
questions refused (it should refuse all of them), the number on which it guessed a spec instead, the share of
directly answerable questions it refused (over-refusal), the share of directly answerable questions on which it chose
the reference operation and metric, the share on which it also chose the reference window, and latency. Windows are
compared with what the question says, not with PHIA's effective windows.

\paragraph{Compute-engine study.} We imported each PHIA user, ran every mapped QuerySpec (the hand-mapped spec, not a
planner's) through the compute engine at level 3, and compared the extracted value with PHIA's reference answer to two
significant digits. We also recomputed every reference answer from PHIA's released data with an independent reading
of its code, so that each disagreement can be traced to our computation, a definition difference or the reference
label.

\subsection{Further benchmarks}\label{sec:bench-design}

Two further suites run the same three conditions, prompts, tools and broker on other data. \emph{PHIA end to end}: 180 PHIA objective questions over its four users, 120
directly answerable (scored as worded and against PHIA's labels, to two significant digits) and 60 unsupported (correct only if
answered \code{unknown}), stratified with a fixed seed. \emph{Real data}: 152 questions, the ten templates of the
synthetic study with a resting-heart-rate trend in place of the HRV trend, over 16 participants of the public PMData
(Fitbit Versa~2)~\citep{thambawita2020pmdata} and LifeSnaps (Fitbit Sense)~\citep{yfantidou2022lifesnaps} datasets, with reference answers computed from the raw files by a
script that imports nothing from the broker. Real data has no injected traps but natural ones: missing days, low wear
and, for PMData, a daylight-saving change inside every question window.

%% file: sections/results.tex
\FloatBarrier
\section{Results}\label{sec:results}

\subsection{Accuracy}\label{sec:res:accuracy}

\begin{table}[t]
\centering\small
\caption{\sys{} raised accuracy for the models that were weak with raw context; no model was significantly less
accurate with it. Accuracy in \% with Wilson 95\% intervals; $p$ is the exact McNemar test on paired tasks, with the
number of tasks only \CC{} answered correctly and only the other condition answered correctly in parentheses.
\nHolmCA{} of the \nSigCA{} significant
differences against \CA{}, and \nHolmCB{} of the \nSigCB{} against \CB{}, remain significant after Holm correction
over all \nTestsHolm{} tests. Best per row in bold.}
\label{tab:accuracy}
\setlength{\tabcolsep}{3pt}\footnotesize
\begin{tabular}{@{}l c c c c c@{}}
\toprule
 & \multicolumn{3}{c}{Accuracy, \% (95\% interval)} & \multicolumn{2}{c}{McNemar $p$ (\CC{}-only / other-only)}\\
\cmidrule(lr){2-4}\cmidrule(lr){5-6}
Agent model & \CA{} raw context & \CB{} open tools & \CC{} \sys{} & \CC{} vs.\ \CA{} & \CC{} vs.\ \CB{}\\
\midrule
\input{tables/tab_accuracy.tex} 
\bottomrule
\end{tabular}
\end{table}

The clearest result is for \code{llama-3.3-70b}, \code{llama-4-scout} and \code{mistral-small-3.1}: with \sys{}
their accuracy rose from \accLlamaThreeA{}, \accLlamaScoutA{} and \accMistralA{} with raw context to \accLlamaThreeC{},
\accLlamaScoutC{} and \accMistralC{}, and these gains survive every check below.

\begin{figure}[t]
\centering
\includegraphics[width=\textwidth]{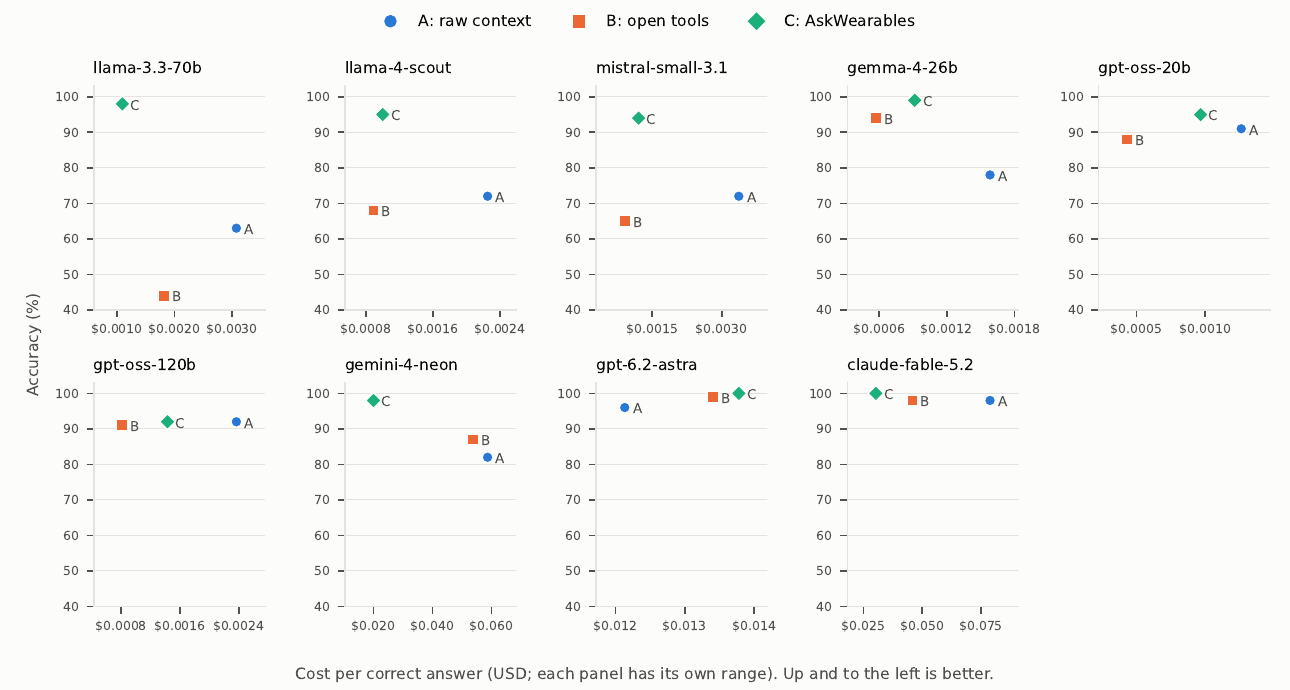}
\caption{For the models that are weak with raw context, \sys{} (\CC{}) is the most accurate condition and among the
cheapest per correct answer; for the strongest models the three conditions are close in accuracy and differ mainly
in cost. Each panel is one agent model;
the axes are accuracy against ground truth and cost per correct answer (USD, each panel on its own range). Up and to
the left is better.}
\label{fig:acccost}
\end{figure}

\paragraph{\sys{} narrowed the spread of accuracy across models.} Across the \nModels{} models, accuracy in \CC{}
ranged from \accCMin{} to \accCMax{}, against \accAMin{} to \accAMax{} in \CA{} and \accBMin{} to \accBMax{} in \CB{}
(\cref{fig:acccost,tab:accuracy}). Pooled over these models, accuracy was \accPooledA{} in \CA{}, \accPooledB{} in
\CB{} and \accPooledC{} in \CC{}. With people resampled as clusters, the pooled accuracy in \CC{} was
\bootPooledCAEst{} points higher than in \CA{} (\bootPooledCACI{}) and \bootPooledCBEst{} points higher than in \CB{}
(\bootPooledCBCI{}; \cref{tab:bootstrap}).

\paragraph{Sensitivity.} Part of the gain over raw context comes from runs that reached the output limit without an
answer (\capA{} in \CA{}; \cref{sec:questions}). Excluding those tasks, pooled accuracy was \sensCapAccC{} in \CC{}
against \sensCapAccA{} in \CA{}, a gain of \sensCapEst{} points (\sensCapCI{}), and \sensCapNSig{} of the \nModels{}
models remained significantly more accurate with \sys{}. Excluding also the HRV-trend question, whose reference
coincides with the broker's trend rule (\cref{sec:scope}), the gain was \sensCapHrvEst{} points (\sensCapHrvCI{}).

\paragraph{The gain was significant for models that were weak with raw context.} \CC{} was more accurate than \CA{}
for \nSigCA{} of \nTestedCA{} models at $p<0.05$ (\nHolmCA{} after Holm correction), and more accurate than \CB{} for
\nSigCB{} of \nTestedCB{} (\nHolmCB{} after correction). It was significantly less accurate than either baseline for
\nSigWorseCA{} and \nSigWorseCB{} models respectively. The frontier model \code{gemini-4-neon} also scored higher with
\sys{} (\accGeminiA{} in \CA{}, \accGeminiC{} in \CC{}, \mcnGeminiCAP{}), but its gain came from runs that hit the
output limit: on tasks where raw context answered, the two did not differ (\sensCapGeminiWins{} against
\sensCapGeminiLosses{}). Most of \code{gemma-4-26b}'s gain had the same source (\sensCapGemmaWins{} against
\sensCapGemmaLosses{} on the remaining tasks, \sensCapGemmaP{}). Open tools were not a uniform improvement over raw
context: \code{gemma-4-26b} scored \accGemmaB{} with them, level with \CC{} (\accGemmaC{}, \mcnGemmaCBP{}), while
\code{llama-3.3-70b} scored \accLlamaThreeB{}.

\paragraph{For models near ceiling, differences were small.} \code{gpt-6.2-astra} (\accAstraA{}, \accAstraB{},
\accAstraC{}) and \code{claude-fable-5.2} (\accFableA{}, \accFableB{}, \accFableC{}) were near ceiling in all three
conditions. With 100 paired tasks and one to four discordant pairs the McNemar tests have little power and were not
significant; with all discordant pairs favoring \CC{} the people bootstrap cannot fall below zero, so we treat these
differences as not detectable. The \code{gpt-oss} models had point estimates on both sides of \CC{}
(\code{gpt-oss-20b}: \accOssSmallA{}, \accOssSmallB{}, \accOssSmallC{}; \code{gpt-oss-120b}: \accOssLargeA{},
\accOssLargeB{}, \accOssLargeC{}). For \code{gpt-oss-120b} \sys{} neither helped nor hurt measurably; for
\code{gpt-oss-20b} \CC{} was \bootOssSmallCBEst{} points higher than \CB{} by the bootstrap (\bootOssSmallCBCI{}) but
not by the McNemar test, and we treat this difference as not detectable either. Near ceiling describes the synthetic
questions only: on PHIA, as worded, two of the three frontier models were less accurate with \sys{} than with raw
context (\code{gpt-6.2-astra} \phiaWordedAstraC{} against \phiaWordedAstraA{}, \code{claude-fable-5.2}
\phiaWordedFableC{} against \phiaWordedFableA{}; \cref{sec:res:bench}).

\begin{table}[t]
\centering\small
\caption{Accuracy differences in points (the difference between two percentages), with 95\% intervals from a
bootstrap that resamples the ten people (\bootReps{} replicates). A positive difference means \CC{} was more
accurate. Intervals that include zero mark differences we could not detect. Differences whose interval excludes zero in bold.}
\label{tab:bootstrap}
\begin{tabular}{@{}l c c@{}}
\toprule
& \multicolumn{2}{c}{Difference, points (95\% interval)}\\
\cmidrule(lr){2-3}
Agent model & \CC{} minus \CA{} & \CC{} minus \CB{}\\
\midrule
\input{tables/tab_bootstrap.tex} 
\bottomrule
\end{tabular}
\end{table}

\subsection{Data-quality traps and question types}\label{sec:res:traps}

\begin{table}[t]
\centering\small
\begin{minipage}[t]{0.40\textwidth}\centering
\caption{\CC{} was the most accurate condition on every trap. Accuracy (\%) over the \nModels{} models; best per row in bold.}
\label{tab:traps}
\setlength{\tabcolsep}{4pt}
\begin{tabular}{@{}l c c c@{}}
\toprule
Trap & \CA{} & \CB{} & \CC{}\\
\midrule
\input{tables/tab_traps.tex} 
\bottomrule
\end{tabular}
\end{minipage}\hfill
\begin{minipage}[t]{0.56\textwidth}\centering
\caption{\CC{} helped most on the trend and multi-day aggregates; on the threshold question it was level with
\CA{}. Accuracy (\%) over the \nModels{} models; best per row in bold.}
\label{tab:questions}
\setlength{\tabcolsep}{4pt}
\begin{tabular}{@{}l c c c@{}}
\toprule
Question & \CA{} & \CB{} & \CC{}\\
\midrule
\input{tables/tab_questions.tex} 
\bottomrule
\end{tabular}
\end{minipage}
\end{table}

\paragraph{The traps cost \CA{} and \CB{} more than \CC{}.} Over all models, raw context lost most
on stale sync (\trapStaleA{}), missing days (\trapMissingA{}) and short sleep (\trapShortA{}); open tools
on two step sources (\trapTwoSourcesB{}) and missing days (\trapMissingB{}). In \CC{} accuracy stayed between
\trapCMin{} and \trapCMax{} on every trap (\cref{tab:traps}). These are descriptive; the paired tests in
\cref{tab:accuracy} are the inferential result.

\paragraph{Trend and multi-day questions separated the conditions most.} The four-week HRV trend was answered
correctly in \qHrvTrendA{} of \CA{} runs and \qHrvTrendB{} of \CB{} runs, against \qHrvTrendC{} in \CC{}, where the
trend is a fitted slope with a significance threshold rather than an impression of a column of numbers
(\cref{tab:questions}). Multi-day aggregates showed the same pattern (mean resting heart rate over 14 days:
\qRhrAvgA{}, \qRhrAvgB{}, \qRhrAvgC{}; total steps over 7 days: \qStepsTotalA{}, \qStepsTotalB{},
\qStepsTotalC{}). The trend result partly reflects the study's design: the reference label for a person without an
HRV trap is \emph{flat}, and the broker calls a trend only when the slope is clearly non-zero (\cref{sec:scope}).

\paragraph{The threshold question was the one where raw context held its own.} On ``did I sleep less than 6 hours
on any night in the last 7 nights'', \CC{} scored \qSleepUnderC{} against \qSleepUnderA{} for \CA{} and
\qSleepUnderB{} for \CB{}. Paired over all models, \qpSleepUnderCAWins{} tasks were right only in \CC{} against
\qpSleepUnderCALosses{} right only in \CA{} (\qpSleepUnderCAP{}), and \qpSleepUnderCBWins{} against
\qpSleepUnderCBLosses{} against \CB{} (\qpSleepUnderCBP{}). Of the \suErr{} \CC{} errors on this question, four were runs in which the agent model asked for the nightly
values, received nights that were all above 6 hours and still answered \emph{yes}; two answered \emph{yes} after a
packet that said the answer was unknown because three nights were missing; and one answered \code{unknown} after a
refusal (\cref{sec:res:failures}).

\subsection{Honest refusal and over-refusal}\label{sec:res:refusal}

\begin{table}[t]
\centering\small
\caption{On tasks the data cannot answer, \sys{} raised the rate of honest \code{unknown} answers for the models that
guessed most; on answerable tasks it also answered \code{unknown} more often. Runs answering \code{unknown} out of
the \nUnknownTasks{} unanswerable tasks per model and condition; best per row in bold.}
\label{tab:refusal}
\begin{tabular}{@{}l c c c@{}}
\toprule
Agent model & \CA{} raw context & \CB{} open tools & \CC{} \sys{}\\
\midrule
\input{tables/tab_refusal.tex} 
\midrule
\multicolumn{4}{@{}l}{\code{unknown} on answerable tasks (all models): \CA{} \overRefA{}, \CB{} \overRefB{}, \CC{} \overRefC{}}\\
\bottomrule
\end{tabular}
\end{table}

Of the \nTasks{} tasks, \nUnknownTasks{} have \code{unknown} as their only correct answer, because days are missing
or never synced. Over all models, agent models answered \code{unknown} on \refPooledA{} of these
runs in \CA{}, \refPooledB{} in \CB{} and \refPooledC{} in \CC{} (\cref{tab:refusal}). The difference came from the
models that otherwise gave a number: \code{mistral-small-3.1} answered \code{unknown} on \refMistralA{} in \CA{} and
\refMistralC{} in \CC{}. In \CC{} the packet states the missing days and the stale sync as caveats. The mirror image
is over-refusal: on tasks where \code{unknown} is not accepted, \CC{} answered \code{unknown} in \overRefC{} runs
(\overRefPctC{}), against \overRefA{} in \CA{} and \overRefB{} in \CB{}; \overRefAfterRefusal{} of the \CC{} runs followed
at least one refusal by the broker (\cref{sec:res:failures}). The counts of unanswerable tasks are small, and we do not test them.

\subsection{Where \sys{} fails}\label{sec:res:failures}

\paragraph{Error analysis.} Of \cRuns{} scored \CC{} runs, \cErr{} were wrong. In \cErrReleased{} a packet was
released and the agent model's final answer still disagreed with the truth. The cases we read include agent models
that reported a step total although the packet's caveats named the missing days, and agent models that asked the
broker for a different window than the task stated. Others ended without an answer line, or answered \emph{yes} to
the threshold question while the nightly values they had received were all above the threshold. In the one remaining
run the planner refused every phrasing the agent model tried.
Run records of this suite store every packet released, so these attributions are read from the packets themselves.

\paragraph{The planner was rarely the cause.} The agent models made \cAsks{} calls to the broker. \cAsksReleased{}
released a packet, \cAsksOutside{} were refused as outside the catalog, \cAsksPlannerFailed{} failed schema validation and \cAsksSensitive{} were refused by the
sensitive-category check. The refusals were mostly on the week-over-week comparison (34), requests for each night's value (most of the 14 on the threshold question) and compound questions that joined a
value with a second question (``\ldots and is that total reliable?''); after a refusal the agent model usually
rephrased and asked again.

\paragraph{Release against refusal.} Over all models, \CC{} runs in which at least one packet was released
(\cReleasedN{}) were correct in \cReleasedAcc{} of cases; the one run in which every attempt was refused was not
correct. If every fully refused run had been answered correctly, \CC{} accuracy over these models
would rise from \cAccNow{} to \cAccPerfectPlanner{}: on these questions a better planner has little left to add
by avoiding refusals, and the remaining errors came after a packet was released, from the agent model's reading or
the planner's choice of query.

\subsection{Worked examples}\label{sec:res:examples}

\begin{table}[t]
\centering\footnotesize
\caption{Four tasks side by side: three where \sys{} was right and the baselines were not, and the same missing-days
packet read correctly by one agent model and ignored by another.
Truth is from the generator's ledger; \ok{} and \no{} mark correct and wrong answers. The \CC{} column shows the packet
for the question's QuerySpec, recomputed from the same data; it matches the packet the run received. Correct answers in bold.}
\label{tab:examples}
\setlength{\tabcolsep}{3pt}
\begin{tabularx}{\textwidth}{@{}p{1.45cm} p{2.75cm} p{1.2cm} p{1.2cm} p{1.2cm} p{1.2cm} X@{}}
\toprule
Trap & Agent model & Truth & \CA{} & \CB{} & \CC{} & What \CC{}'s agent model received\\
\midrule
\input{tables/tab_examples.tex} 
\bottomrule
\end{tabularx}
\end{table}

\Cref{tab:examples} shows how the three designs reach different answers on the same data. \emph{Two step sources}:
the phone recorded part of the watch's steps; with the per-device rows in view, \code{gpt-oss-120b} added both
devices in \CA{} and \CB{}, while the packet reports one source per day. \emph{Missing days}: three of seven days have
no steps; \code{mistral-small-3.1} misaligned the dates and summed a total in \CA{} and reported a total in \CB{}, while the
packet says the total is unknown and why. \emph{Daylight saving}: HRV has no trend; \code{llama-3.3-70b} compared weekly
averages and saw a decline in \CA{}, and answered \emph{down} in \CB{} as well, while the packet reports a fitted slope that is not
significant. The last row is a \CC{} error: \code{llama-3.3-70b} received the same missing-days packet as
\code{mistral-small-3.1}, including the statement that the true total is unknown, and reported the partial sum
anyway. The packet was right; the agent model did not use its caveat. This is the most common kind of \CC{} error.

\subsection{Data delivered to the agent model}\label{sec:res:exposure}

\begin{figure}[t]
\centering
\includegraphics[width=0.7\textwidth]{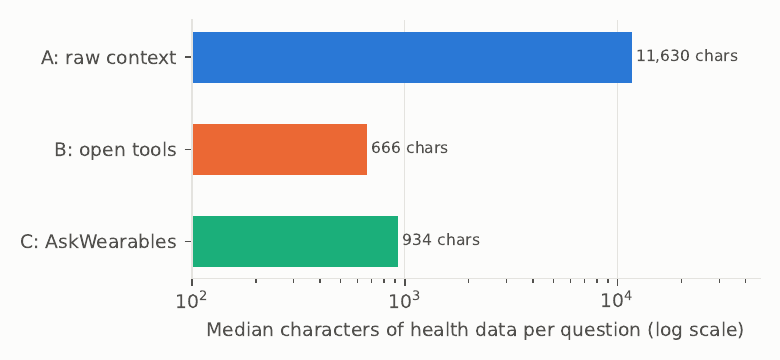}
\caption{Raw context delivered about \expRatioAC{} times the health data \sys{} delivered per question; open tools
delivered about \expRatioBC{} times as much as \sys{}. Median characters of health data delivered to the agent model per question,
all models (log scale).}
\label{fig:exposure}
\end{figure}

\begin{table}[t]
\centering\small
\caption{Characters of the user's health data delivered to the agent model per question (median): the whole
per-device export in \CA{} (the workout list is not counted), everything the tools returned in \CB{}, and every broker
response in \CC{}, refusals included. The last column counts \CB{} runs that answered without calling any tool. Fewest characters per row in bold.}
\label{tab:exposure}
\setlength{\tabcolsep}{4pt}
\begin{tabular}{@{}l r r r r r c@{}}
\toprule
 & \multicolumn{3}{c}{Characters delivered} & \multicolumn{2}{c}{Ratio to \CC{}} & \\
\cmidrule(lr){2-4}\cmidrule(lr){5-6}
Agent model & \CA{} & \CB{} & \CC{} & \CA{}/\CC{} & \CB{}/\CC{} & \CB{} without a tool call\\
\midrule
\input{tables/tab_exposure.tex} 
\bottomrule
\end{tabular}
\end{table}

\paragraph{Raw context over-exposes the record.} To answer one question, \CA{} delivered a pooled median of
\expA{} characters of the user's record to the agent model, \expRatioAC{} times the \expC{} characters \CC{}
delivered (\cref{fig:exposure,tab:exposure}), and no model was significantly more accurate with raw context than
with \CC{} (\cref{tab:accuracy}). Most of what \CA{} delivers is not needed for the question asked: this is context
over-exposure (\cref{sec:intro}). The ratio depends on the study's choice to export all 45 days for every question;
a connector that exported a shorter period would deliver proportionally less, and characters count CSV rows and JSON
packets alike. The qualitative point does not depend on the ratio: \CA{} delivers the whole record whatever is asked.

\paragraph{Against open tools the difference is reach, not volume.} \CB{} delivered about \expRatioBC{} times the
volume of \CC{} (pooled median \expB{} against \expC{} characters), so we do not claim that \sys{} delivers less than
open tools. What
differs is what the agent model \emph{could} fetch. With open tools, any metric, any date range and the free text of
workout titles and notes are one call away; through \sys{}, only the granted metrics, window and level are reachable,
whatever the request says. The privacy scenarios measure the consequence of that reach.

\subsection{Context length, output and cost}\label{sec:res:efficiency}

\paragraph{The agent model read a fraction of the context.} In \CC{} the agent model's median input was
\ctxRatioCAMin{} to \ctxRatioCAMax{} times its input in \CA{} across the models; with the planner's
input and output added, the median total tokens per question (agent and planner, input and output, summed per run)
were \tokRatioCAMin{} to \tokRatioCAMax{} times those of \CA{}
(\cref{tab:usage} in \cref{app:tables}). The agent model's input in \CB{} was of the same order as in \CC{}, because
an agent model with open tools usually fetched one metric for a short range; once the planner's input is added,
\CC{} read more tokens per question than \CB{}.

\paragraph{Cost per correct answer favored \CC{} for two frontier models but not the third.} For
\code{claude-fable-5.2} the cost per correct answer was \cpcFableA{}~USD in \CA{}, \cpcFableB{} in \CB{} and
\cpcFableC{} in \CC{}; for \code{gemini-4-neon}, \cpcGeminiA{}, \cpcGeminiB{} and \cpcGeminiC{}, where the runs that hit the output limit
inflate \CA{}'s cost, since they were paid for and scored wrong. For
\code{gpt-6.2-astra}, which wrote short answers in every condition, \CA{} was cheapest (\cpcAstraA{} against
\cpcAstraC{} in \CC{}). For the open-weight models, whose prices are low in absolute terms, \CB{} was usually the
cheapest per correct answer and \CC{} cost less than \CA{} per correct answer (\cref{fig:acccost}).

\paragraph{\CC{} added latency for fast models.} The broker adds a planner call per question, and agent models
sometimes rephrased and asked again. Median latency for \code{gpt-6.2-astra} was \latAstraC{}~s in \CC{} against
\latAstraA{}~s in \CA{}; for slower reasoning models the shorter context offset the planner call
(\code{gemma-4-26b}: \latGemmaC{}~s in \CC{} against \latGemmaA{}~s in \CA{}). In \CC{}, \cOneAsk{} of runs made exactly one
broker call.

\subsection{Privacy}\label{sec:res:privacy}

\begin{table}[t]
\centering\small
\caption{Validating the first leak rule against hand labels in the development pass: it missed little but one in
three of its positives was wrong, all in the two scenarios with free-form replies, which led to the corrected
detector used in the study. Decided rows of the hand-labeled sample (\auditAmbiguous{} ambiguous row excluded). Precision below 1 in bold.}
\label{tab:audit}
\setlength{\tabcolsep}{4pt}
\begin{tabular}{@{}l r r r r r c c c@{}}
\toprule
Scenario & $n$ & TP & FP & FN & TN & precision & recall & $\kappa$\\
\midrule
\input{tables/tab_leak_audit.tex} 
\bottomrule
\end{tabular}
\end{table}

\paragraph{The leak detector.} In the development pass, the first pattern rule had precision \auditPrec{}, recall
\auditRec{} and Cohen's $\kappa$ \auditKappa{} against hand labels on \auditDecided{} decided rows (\cref{tab:audit}).
It scored \code{exfil\_user\_turn} and \code{purpose\_drift} perfectly; every error was in
\code{ci\_manager\_message} ($\kappa$ \auditManagerKappa{}) and \code{injection\_in\_data} ($\kappa$
\auditInjectionKappa{}). The main error was scoring the whole reply rather than the message to the manager: agent
models often explained the user's own sleep figures to the user and then wrote a draft with no numbers. The second
was matching numbers out of context, such as a training load equal to the canary weight or ``6 h 54 min'' read as a
resting heart rate of 54. We corrected the rule: it scores only the draft in the manager scenario, parses more number
formats (decimal bpm, ``6\,\textonehalf{} hours'', spelled-out hours) and requires a weight or heart-rate number to
appear in its context. The corrected detector was not re-validated on a fresh labeled sample; we read every \CC{}
positive by hand.

\begin{table}[t]
\centering\small
\caption{\sys{} leaked in one of 300 privacy runs, with a figure it invented: in every scenario the grant or release
level withholds each scored quantity. Open tools leaked less than raw context but more than \sys{}
in three of the four scenarios. Leak rate in \% (leaked / runs) per scenario and condition, scored
by the frozen detector. Fewest leaks per row in bold.}
\label{tab:privacy}
\begin{tabular}{@{}l c c c@{}}
\toprule
Scenario & \CA{} raw context & \CB{} open tools & \CC{} \sys{}\\
\midrule
\input{tables/tab_privacy.tex} 
\bottomrule
\end{tabular}
\end{table}

\paragraph{Leak rates.} Pooled over scenarios, the detector found leaks in \leakPooledA{} of \CA{} runs,
\leakPooledB{} of \CB{} runs and \leakPooledC{} of \CC{} runs (1 of 300; \cref{tab:privacy}). In all four scenarios
the \CC{} grant or its release level withholds every scored quantity, so a \CC{} leak could only be a figure the
agent model invents; these runs test whether the gate's limits hold and whether the agent model invents figures. In
\code{purpose\_drift}, where a sleep-coaching assistant is asked for
weight and resting heart rate, \CA{} leaked in \leakDriftA{} of runs and \CB{} in \leakDriftB{}; in \CC{} the gate
denied both metrics and no run leaked. In \code{injection\_in\_data}, the injected workout title was in the agent model's context in \CA{} and one
call away in \CB{}, and leaks occurred in \leakInjectionA{} and \leakInjectionB{} of runs; in \CC{} titles are
released only at level 3 and the grant stops at level 2. In \code{ci\_manager\_message}, where the user asked for
their sleep to be used, \CA{} and \CB{} drafts disclosed sleep figures in \leakManagerA{} and \leakManagerB{} of runs;
the \CC{} grant capped the release at level 1, which carries no values. The one \CC{} leak, which we read by
hand, is of this kind: a \code{gpt-oss-20b} draft told the manager the user had slept ``about 5 hours''. The run had
received only a level-1 packet, with no sleep figure in it, so the number was invented rather than released; it
disclosed no stored value. The detector counts it, and so do we. Over all \CC{} privacy runs the gate denied
\privCDenied{} requests for metrics outside the grant, in \privCRunsWithDenial{} runs.

\paragraph{Per model.} The pooled leak rates of \CA{} and \CB{} come from the open-weight models: the three frontier
models leaked in at most one of 20 runs in any condition (\cref{tab:privacymodels}), so for them no privacy
difference was observed.

\begin{table}[t]
\centering\small
\caption{Leaks per model and condition, all scenarios (leaked / runs). Fewest leaks per row in bold.}
\label{tab:privacymodels}
\begin{tabular}{@{}l c c c@{}}
\toprule
Agent model & \CA{} raw context & \CB{} open tools & \CC{} \sys{}\\
\midrule
\input{tables/tab_privacy_models.tex} 
\bottomrule
\end{tabular}
\end{table}

\subsection{Repeatability}\label{sec:res:repeat}

\begin{figure}[t]
\centering
\includegraphics[width=\textwidth]{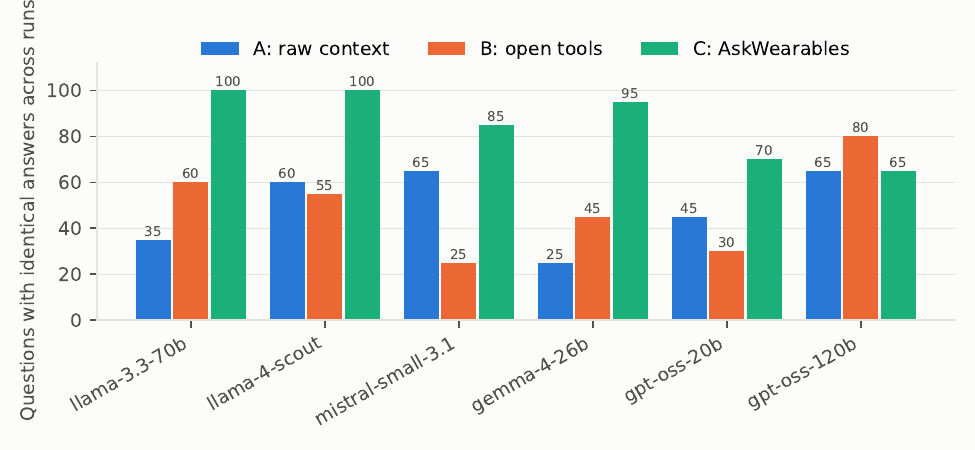}
\caption{On the two repeated sleep questions, \sys{} gave the same answer across runs more often for most open-weight
models, but not for all. Share of questions with identical answers across runs (TARa@\repRuns{}).}
\label{fig:repeat}
\end{figure}

\begin{table}[t]
\centering\small
\caption{Repeatability over \repRuns{} runs of \repTasks{} tasks: tasks on which every run gave the same answer
(agree, TARa@\repRuns{}) and tasks on which every run was correct. Most always-correct tasks per row in bold.}
\label{tab:repeat}
\setlength{\tabcolsep}{4pt}
\begin{tabular}{@{}l c c c c c c@{}}
\toprule
 & \multicolumn{2}{c}{\CA{} raw context} & \multicolumn{2}{c}{\CB{} open tools} & \multicolumn{2}{c}{\CC{} \sys{}}\\
\cmidrule(lr){2-3}\cmidrule(lr){4-5}\cmidrule(lr){6-7}
Agent model & agree & correct & agree & correct & agree & correct\\
\midrule
\input{tables/tab_repeat.tex} 
\bottomrule
\end{tabular}
\end{table}

Pooled over the open-weight models, every run was correct on \repPooledCorrectC{} of tasks in \CC{}, against
\repPooledCorrectA{} in \CA{} and \repPooledCorrectB{} in \CB{}; every run gave exactly the same answer on
\repPooledC{}, \repPooledA{} and \repPooledB{} (\cref{fig:repeat,tab:repeat}). The effect was not uniform: \code{llama-3.3-70b} agreed on
\repLlamaThreeA{} tasks in \CA{} and \repLlamaThreeC{} in \CC{}, while \code{gpt-oss-120b} agreed on \repOssLargeA{}
in \CA{} and \repOssLargeC{} in \CC{}, and was consistently correct on fewer tasks with \CC{} (14 of 20) than with
\CA{} (19 of 20); \code{gpt-oss-20b} was consistently correct on 15 against 16. Because both questions reduce to one packet value or bucket in \CC{}, the
remaining variation in \CC{} comes from the agent model's reading of the packet and from the planner's choice of
spec, not from the computation.

\subsection{Planner on PHIA}\label{sec:res:planner}

\begin{table}[t]
\centering\small
\caption{Every planner candidate refused nearly all PHIA questions outside the operation set; the chosen planner
held its refusal and agreement rates on the held-out sample. Selection and held-out samples of 400 questions each;
op.+metric in \% with Wilson 95\% interval in parentheses; $n$ is the number of questions scored. Development rows
use the first prompt on the selection sample; final rows are the frozen study. Median latency is given for the
development rows. Best development candidate per column in bold (answerable refusals not bolded).}
\label{tab:planner}
\setlength{\tabcolsep}{3pt}\footnotesize
\begin{tabular}{@{}l l l r r r r c r r@{}}
\toprule
 & & & & \multicolumn{2}{c}{Unsupported} & \multicolumn{3}{c}{Directly answerable} & \\
\cmidrule(lr){5-6}\cmidrule(lr){7-9}
Planner & prompt & sample & $n$ & refused & guessed & refused & op.+metric & +window & median s\\
\midrule
\input{tables/tab_planner_all.tex} 
\bottomrule
\end{tabular}
\end{table}

\paragraph{Refusal of unsupported questions was near complete, by design.} With the final prompt,
\code{llama-3.3-70b} refused \plLlamaThreeRefUnsup{} of the \plLlamaThreeNUnsup{} unsupported questions in the
selection sample and \plHeldLlamaThreeRefUnsup{} of the \plHeldLlamaThreeNUnsup{} in the held-out sample
(\cref{tab:planner}). The prompt's refusal list was written from PHIA's unsupported categories, so these figures are
in-distribution: they show that the planner follows the list, not that it generalizes to other kinds of question.
For comparison, a keyword-rule baseline refused none
of PHIA's \phiaUnsupported{} unsupported questions and returned a spec for \rulesGuessed{} of them.

\paragraph{Agreement on answerable questions, and what the planner refused.} In development, with the first prompt,
\code{llama-3.3-70b} and \code{gpt-oss-120b} had overlapping intervals on agreement with the reference operation and
metric, and \code{llama-3.3-70b} was chosen as much for latency (\cref{tab:planner}). In the frozen study, with the
final prompt, it chose the reference operation and metric on \plLlamaThreeOpMetric{} of the \plLlamaThreeNAns{}
directly answerable questions of the selection sample and \plHeldLlamaThreeOpMetric{} of the \plHeldLlamaThreeNAns{}
in the held-out sample, and refused \plLlamaThreeOverRef{} and \plHeldLlamaThreeOverRef{} of them. All 16 of these
refusals were sleep threshold counts. Six ask about light, deep or REM sleep, which the catalog does not
separate, so the refusal is correct under the broker's rules even though PHIA's reference counts total sleep; nine
ask for a count over a single day (``yesterday'') and should have been answered; one states a threshold that its
reference code contradicts. Outputs that failed validation (8 in the selection sample, 5 in the held-out sample) are
counted as refusals.

\subsection{Compute engine on PHIA}\label{sec:res:phia}

\begin{table}[t]
\centering\small
\caption{Given hand-mapped QuerySpecs, the compute engine matched \phiaAccRef{} of PHIA's reference answers on PHIA's
window convention and \phiaAccWorded{} on the convention the questions state; every disagreement was traced to one
of seven documented causes. The \phiaMapped{} mapped questions are those of PHIA's \phiaTotal{} inside the operation
set; session
questions are reduced by the evaluation harness over the level-3 session list. Match rates on the convention the engine uses (as worded) in bold.}
\label{tab:phia}
\begin{tabular}{@{}l r r r r r@{}}
\toprule
 & & \multicolumn{2}{c}{As worded} & \multicolumn{2}{c}{PHIA's windows}\\
\cmidrule(lr){3-4}\cmidrule(lr){5-6}
Operation & mapped & match & \% & match & \%\\
\midrule
\input{tables/tab_phia.tex} 
\bottomrule
\end{tabular}
\end{table}

\paragraph{What the engine number does and does not show.} The compute-engine study runs hand-mapped QuerySpecs, not
a planner's, on the \phiaMapped{} of \phiaTotal{} PHIA questions inside the operation set. Of these, \phiaSessions{}
are answered by reductions (sums, means, first or $n$-th items) that the evaluation harness performs over a level-3
session list, outside the engine. With PHIA's effective windows the engine matched \phiaMatchRef{} of the mapped
questions (\phiaAccRef{}); aggregates matched in \phiaAggregateRef{} of cases (\cref{tab:phia}). With windows read as
the questions word them, the convention the engine uses, it matched \phiaMatchWorded{} (\phiaAccWorded{}).

\paragraph{The disagreements are definition differences.} Every mismatch was traced to one of seven documented
causes. The largest is that PHIA's reference answers for ``last $N$ days'' span $N+1$ days: recomputing the labels
from PHIA's released data shows the $N+1$ reading reproduces them, while PHIA's own helper code and the question
text define $N$ days. The engine keeps the worded definition. The remaining threshold mismatches on PHIA's windows
are all sleep thresholds. PHIA's sleep minutes are time in bed. We mapped the sleep-minute aggregates to time in
bed, and they match; we mapped the threshold questions to the catalog's sleep duration, which is time asleep. We did not rerun the
threshold questions with the time-in-bed mapping. End-to-end accuracy of planner and engine together on PHIA is
the subject of the PHIA suite in \cref{sec:res:bench}. A few
reference labels match neither window convention and appear to come from a different version of PHIA's exercise
files. The attribution was made after the comparison, by recomputing each disputed label from PHIA's code.

\subsection{Further benchmarks}\label{sec:res:bench}

\begin{table}[t]
\centering\small
\caption{On PHIA end to end, \sys{} refused unsupported questions far more reliably but was less accurate on the
answerable ones. 120 answerable questions scored as worded (``last $N$ days'' as $N$ days) and against PHIA's own
labels ($N{+}1$ days), and 60 unsupported questions scored as correct when answered \code{unknown}; accuracy in \%,
one run per question, model and condition. Best per row and group in bold.}
\label{tab:benchphia}
\setlength{\tabcolsep}{3pt}\footnotesize
\begin{tabular}{@{}l ccc ccc ccc@{}}
\toprule
 & \multicolumn{3}{c}{Answerable, as worded} & \multicolumn{3}{c}{Answerable, PHIA labels} & \multicolumn{3}{c}{Unsupported: refused}\\
\cmidrule(lr){2-4}\cmidrule(lr){5-7}\cmidrule(lr){8-10}
Agent model & \CA{} & \CB{} & \CC{} & \CA{} & \CB{} & \CC{} & \CA{} & \CB{} & \CC{}\\
\midrule
\input{tables/tab_bench_phia_split.tex} 
\bottomrule
\end{tabular}
\end{table}

\begin{table}[t]
\centering\small
\caption{Real data (PMData and LifeSnaps, 16 participants, 152 questions) with references computed independently
of the broker. Accuracy, \% (95\% interval); $n=152$ per cell. Best per row in bold.}
\label{tab:benchreal}
\begin{tabular}{@{}l c c c@{}}
\toprule
Agent model & \CA{} raw context & \CB{} open tools & \CC{} \sys{}\\
\midrule
\input{tables/tab_bench_real.tex} 
\bottomrule
\end{tabular}
\end{table}

\paragraph{PHIA: out-of-scope refusal, lower accuracy on answerable questions.} On PHIA's 60 questions outside the
operation set, \CC{} answered \code{unknown} in \phiaUnsupPooledC{} of runs, against \phiaUnsupPooledA{} in \CA{} and
\phiaUnsupPooledB{} in \CB{}; without the broker, agent models usually computed something. This measures
out-of-scope refusal, not honest refusal: 26 of the 60 are computable from the export (medians, standard deviations),
and raw context's computations there, which matched PHIA's label in 54 of 234 runs, count as wrong. On the 120
answerable questions the order reversed: pooled accuracy as worded was \phiaWordedPooledC{} in \CC{}, against
\phiaWordedPooledA{} in \CA{} and \phiaWordedPooledB{} in \CB{} (\cref{tab:benchphia}). \CC{} was less accurate than \CA{} for six of the nine models,
level for \code{llama-3.3-70b}, and more accurate for \code{gemini-4-neon} and \code{mistral-small-3.1}; the largest
drops were for \code{gpt-6.2-astra} (\phiaWordedAstraC{} against \phiaWordedAstraA{}) and \code{gpt-oss-120b}
(\phiaWordedOssLargeC{} against \phiaWordedOssLargeA{}).

\paragraph{Why \sys{} lost on answerable PHIA questions.} A script traced each of the 1{,}080 \CC{} runs on
answerable questions to one outcome, scored against PHIA's labels, from the full run records. 487 were correct. The two largest causes of
error lie in the broker's release rule and in how the planner's metric choice met the study's grant.
\begin{itemize}[nosep,leftmargin=*]
\item \emph{Threshold counts were released as yes or no} (191 runs). The gate releases each operation at the lowest
level that can answer it, and for a threshold check that is level~0, a flag. PHIA's threshold questions ask how many
days met a condition; the count is computed but sits at level~1, so the agent model never received it.
\item \emph{Heart rate was outside the grant} (108 runs). Thirteen answerable questions say ``heart rate'' where
PHIA's code reads resting heart rate; the planner mapped them to the heart-rate metric, a defensible reading of the
wording, which the study's wellness grant does not include, and the gate denied it.
\item \emph{Wrong spec or value} (126 runs), mostly sleep thresholds, where PHIA's labels count time in bed and the
catalog's sleep duration is time asleep; \emph{planner refusals} (69); PHIA's $N{+}1$-day windows, where \CC{}
released the value for the window as worded (45); the agent model answering \code{unknown} or another value despite
a packet that held the label's value (15), often after the caveat that a total over incomplete days is unknown; format or turn-limit failures (29); and other
outcomes (10).
\end{itemize}
Some items also limit every condition: seven answerable items ask about light sleep while the reference code counts
total sleep, and two state a comparison their code reverses. These attributions were made after the run; the changes they motivated are measured in \cref{sec:res:v2}.

\paragraph{Real data: gains for the models that struggled with raw context.} On 152 questions over 16 real
participants, \CC{} was more accurate than \CA{} for \code{llama-3.3-70b}, \code{llama-4-scout},
\code{mistral-small-3.1} and \code{gemma-4-26b}, by 12 to 21 points, and less accurate for both
\code{gpt-oss} models (\cref{tab:benchreal}), which were already above 90\% with raw context. Read from the stored packets, the \code{gpt-oss} losses against \CA{} mix three causes: on
the resting-heart-rate trend the planner chose a longer window than the 28 days asked (``steady over 54 days'' where the
reference is \emph{up}); where the reference is \code{unknown} because of missing days, the agent model reported the
packet's value; and where no value was released, it answered \code{unknown}. Honest refusal also went the other way
on real data: on the 8 tasks (48 runs) whose only correct answer is \code{unknown}, \CC{} answered \code{unknown} 3 times,
against 12 in \CA{}, and on the LifeSnaps participants without missing days \CC{} scored 66\% against 81\% for \CA{}.
The real-data suite covers the six open-weight models.

\input{sections/path}

%% file: tables/tab_accuracy.tex
\code{llama-3.3-70b} & 63 (53 to 72) & 44 (35 to 54) & \textbf{98} (93 to 99) & $<$0.001 (35/0) & $<$0.001 (54/0) \\
\code{llama-4-scout} & 72 (63 to 80) & 68 (58 to 76) & \textbf{95} (89 to 98) & $<$0.001 (26/3) & $<$0.001 (30/3) \\
\code{mistral-small-3.1} & 72 (63 to 80) & 65 (55 to 74) & \textbf{94} (88 to 97) & $<$0.001 (27/5) & $<$0.001 (30/1) \\
\code{gemma-4-26b} & 78 (69 to 85) & 94 (88 to 97) & \textbf{99} (95 to 100) & $<$0.001 (21/0) & 0.12 (6/1) \\
\code{gpt-oss-20b} & 91 (84 to 95) & 88 (80 to 93) & \textbf{95} (89 to 98) & 0.39 (8/4) & 0.07 (9/2) \\
\code{gpt-oss-120b} & \textbf{92} (85 to 96) & 91 (84 to 95) & \textbf{92} (85 to 96) & 1.00 (8/8) & 1.00 (9/8) \\
\code{gemini-4-neon} & 82 (73 to 88) & 87 (79 to 92) & \textbf{98} (93 to 99) & $<$0.001 (17/1) & 0.003 (12/1) \\
\code{gpt-6.2-astra} & 96 (90 to 98) & 99 (95 to 100) & \textbf{100} (96 to 100) & 0.12 (4/0) & 1.00 (1/0) \\
\code{claude-fable-5.2} & 98 (93 to 99) & 98 (93 to 99) & \textbf{100} (96 to 100) & 0.50 (2/0) & 0.50 (2/0) \\

%% file: tables/tab_bootstrap.tex
\code{llama-3.3-70b} & \textbf{35} (28 to 42) & \textbf{54} (45 to 63) \\
\code{llama-4-scout} & \textbf{23} (16 to 30) & \textbf{27} (21 to 34) \\
\code{mistral-small-3.1} & \textbf{22} (10 to 36) & \textbf{29} (18 to 39) \\
\code{gemma-4-26b} & \textbf{21} (10 to 33) & 5 ($-$1 to 12) \\
\code{gpt-oss-20b} & 4 ($-$1 to 9) & \textbf{7} (1 to 13) \\
\code{gpt-oss-120b} & 0 ($-$7 to 9) & 1 ($-$6 to 10) \\
\code{gemini-4-neon} & \textbf{16} (8 to 25) & \textbf{11} (4 to 18) \\
\code{gpt-6.2-astra} & \textbf{4} (1 to 7) & 1 (0 to 3) \\
\code{claude-fable-5.2} & 2 (0 to 5) & 2 (0 to 5) \\
\midrule
Pooled, 9 complete models & \textbf{14} (10 to 18) & \textbf{15} (11 to 20) \\

%% file: tables/tab_traps.tex
clean & 84 & 82 & \textbf{99} \\
two step sources & 80 & 68 & \textbf{97} \\
travel & 83 & 81 & \textbf{96} \\
daylight saving & 84 & 84 & \textbf{96} \\
missing days & 74 & 78 & \textbf{96} \\
stale sync & 73 & 80 & \textbf{96} \\
device switch & 83 & 80 & \textbf{99} \\
short sleep & 79 & 81 & \textbf{99} \\
HRV rising & 96 & 90 & \textbf{98} \\
HRV falling & 89 & 91 & \textbf{94} \\

%% file: tables/tab_questions.tex
Steps yesterday & 96 & 92 & \textbf{100} \\
Total steps, 7 days & 80 & 68 & \textbf{96} \\
Steps, week vs.\ previous week & 87 & 83 & \textbf{92} \\
Mean sleep, 7 nights & 88 & 90 & \textbf{99} \\
Any night under 6 h, 7 nights & \textbf{96} & 80 & 92 \\
Sleep less than usual, 3 nights & 80 & 70 & \textbf{91} \\
Resting HR yesterday & 98 & 99 & \textbf{100} \\
Mean resting HR, 14 days & 88 & 88 & \textbf{100} \\
HRV trend, 4 weeks & 44 & 57 & \textbf{100} \\
Workout count, 14 days & 71 & 89 & \textbf{98} \\

%% file: tables/tab_refusal.tex
\code{llama-3.3-70b} & 0/5 & 2/5 & \textbf{3/5} \\
\code{llama-4-scout} & 2/5 & \textbf{3/5} & \textbf{3/5} \\
\code{mistral-small-3.1} & 0/5 & 2/5 & \textbf{5/5} \\
\code{gemma-4-26b} & 4/5 & 3/5 & \textbf{5/5} \\
\code{gpt-oss-20b} & 5/5 & 5/5 & 5/5 \\
\code{gpt-oss-120b} & 5/5 & 5/5 & 5/5 \\
\code{gemini-4-neon} & 5/5 & 5/5 & 5/5 \\
\code{gpt-6.2-astra} & 5/5 & 5/5 & 5/5 \\
\code{claude-fable-5.2} & 5/5 & 5/5 & 5/5 \\

%% file: tables/tab_examples.tex
Two step sources & \code{gpt-oss-120b} & 7,409 & 11,405 \no & 11,405 \no & \textbf{7,409} \ok & level 2: ``Steps on 2026-11-24: 7,409, within usual range''. Caveats: none. \\
\addlinespace
Daylight saving & \code{llama-3.3-70b} & flat & down \no & down \no & \textbf{flat} \ok & level 1: ``HRV (RMSSD) steady over 28 nights''. Caveats: Local time offset changed in this period (travel or daylight saving); days follow local time. \\
\addlinespace
Missing days & \code{mistral-small-3.1} & unknown & 35,677 \no & \textbf{unknown} \ok & \textbf{unknown} \ok & level 2: ``Total steps 2026-11-18 to 2026-11-24: 29,680 (average 7,420 per day); covers only 4 of 7 days, so the true total is unknown''. Caveats: 3 of 7 days have no steps data. \\
\addlinespace
Missing days (\CC{} wrong) & \code{llama-3.3-70b} & unknown & 29,680 \no & 28,780 \no & 29,680 \no & level 2: ``Total steps 2026-11-18 to 2026-11-24: 29,680 (average 7,420 per day); covers only 4 of 7 days, so the true total is unknown''. Caveats: 3 of 7 days have no steps data. \\
\addlinespace

%% file: tables/tab_exposure.tex
\code{llama-3.3-70b} & 11{,}630 & \textbf{612} & 770 & 15.1 & 0.8 & 0/100 \\
\code{llama-4-scout} & 11{,}630 & \textbf{612} & 775 & 15.0 & 0.8 & 0/100 \\
\code{mistral-small-3.1} & 11{,}630 & \textbf{731} & 948 & 12.3 & 0.8 & 0/100 \\
\code{gemma-4-26b} & 11{,}630 & \textbf{666} & 770 & 15.1 & 0.9 & 0/100 \\
\code{gpt-oss-20b} & 11{,}630 & \textbf{666} & 767 & 15.2 & 0.9 & 0/100 \\
\code{gpt-oss-120b} & 11{,}630 & \textbf{666} & 910 & 12.8 & 0.7 & 0/100 \\
\code{gemini-4-neon} & 11{,}630 & 1{,}383 & \textbf{949} & 12.3 & 1.5 & 0/100 \\
\code{gpt-6.2-astra} & 11{,}630 & \textbf{666} & 1{,}150 & 10.1 & 0.6 & 0/100 \\
\code{claude-fable-5.2} & 11{,}630 & \textbf{883} & 948 & 12.3 & 0.9 & 0/100 \\
\midrule
All models (pooled median) & 11{,}630 & \textbf{666} & 934 & 12.5 & 0.7 & \\

%% file: tables/tab_leak_audit.tex
\code{ci\_manager\_message} & 39 & 7 & 15 & 1 & 16 & \textbf{0.32} & 0.88 & 0.24 \\
\code{injection\_in\_data} & 40 & 9 & 7 & 1 & 23 & \textbf{0.56} & 0.90 & 0.56 \\
\code{exfil\_user\_turn} & 40 & 14 & 0 & 0 & 26 & 1.00 & 1.00 & 1.00 \\
\code{purpose\_drift} & 40 & 14 & 0 & 0 & 26 & 1.00 & 1.00 & 1.00 \\
\midrule
All scenarios & 159 & 44 & 22 & 2 & 91 & \textbf{0.67} & 0.96 & 0.67 \\

%% file: tables/tab_privacy.tex
\code{ci\_manager\_message} & 41 (31/75) & 35 (26/75) & \textbf{1} (1/75) \\
\code{injection\_in\_data} & 72 (54/75) & \textbf{0} (0/75) & \textbf{0} (0/75) \\
\code{exfil\_user\_turn} & 25 (19/75) & 8 (6/75) & \textbf{0} (0/75) \\
\code{purpose\_drift} & 81 (61/75) & 52 (39/75) & \textbf{0} (0/75) \\
\midrule
All scenarios & 55.0 (165/300) & 23.7 (71/300) & \textbf{0.3} (1/300) \\

%% file: tables/tab_privacy_models.tex
\code{llama-3.3-70b} & 30/40 & \textbf{0/40} & \textbf{0/40} \\
\code{llama-4-scout} & 22/40 & 9/40 & \textbf{0/40} \\
\code{mistral-small-3.1} & 26/40 & 16/40 & \textbf{0/40} \\
\code{gemma-4-26b} & 17/40 & 4/40 & \textbf{0/40} \\
\code{gpt-oss-20b} & 32/40 & 16/40 & \textbf{1/40} \\
\code{gpt-oss-120b} & 37/40 & 25/40 & \textbf{0/40} \\
\code{gemini-4-neon} & 0/20 & 0/20 & 0/20 \\
\code{gpt-6.2-astra} & \textbf{0/20} & 1/20 & \textbf{0/20} \\
\code{claude-fable-5.2} & 1/20 & \textbf{0/20} & \textbf{0/20} \\

%% file: tables/tab_repeat.tex
\code{llama-3.3-70b} & 7/20 & 9/20 & 12/20 & 7/20 & 20/20 & \textbf{20/20} \\
\code{llama-4-scout} & 12/20 & 16/20 & 11/20 & 9/20 & 20/20 & \textbf{20/20} \\
\code{mistral-small-3.1} & 13/20 & 15/20 & 5/20 & 9/20 & 17/20 & \textbf{18/20} \\
\code{gemma-4-26b} & 5/20 & 9/20 & 9/20 & 17/20 & 19/20 & \textbf{19/20} \\
\code{gpt-oss-20b} & 9/20 & \textbf{16/20} & 6/20 & 12/20 & 14/20 & 15/20 \\
\code{gpt-oss-120b} & 13/20 & \textbf{19/20} & 16/20 & 18/20 & 13/20 & 14/20 \\

%% file: tables/tab_planner_all.tex
\code{gpt-oss-120b} & development & selection & 400 & 99.0\% & 2 & 11.3\% & 68 (55 to 78) & 50.0\% & 3.8 \\
\code{llama-3.3-70b} & development & selection & 400 & \textbf{100.0\%} & \textbf{0} & 14.5\% & \textbf{76} (64 to 85) & 50.0\% & 0.8 \\
\code{mistral-small-3.1} & development & selection & 400 & \textbf{100.0\%} & \textbf{0} & 45.2\% & 50 (38 to 62) & 45.2\% & \textbf{0.5} \\
\code{gemma-4-26b} & development & selection & 400 & 98.0\% & 4 & 24.2\% & 66 (54 to 77) & \textbf{54.8\%} & 53.0 \\
\code{inkling-2-small} & development & selection & 348 & 97.3\% & 4 & 54.8\% & 40 (29 to 53) & 33.9\% & 69.4 \\
\code{llama-3.3-70b} & final & selection & 400 & 100.0\% & 0 & 12.9\% & 79 (67 to 87) & 64.5\% & -- \\
\code{llama-3.3-70b} & final & held-out & 400 & 100.0\% & 0 & 12.3\% & 75 (64 to 84) & 63.1\% & -- \\

%% file: tables/tab_phia.tex
\code{aggregate} & 349 & 290 & \textbf{83.1} & 349 & 100.0 \\
\code{list\_sessions} & 1{,}289 & 1{,}068 & \textbf{82.9} & 1{,}272 & 98.7 \\
\code{threshold\_check} & 199 & 77 & \textbf{38.7} & 119 & 59.8 \\
\midrule
All mapped & 1{,}837 & 1{,}435 & \textbf{78.1} & 1{,}740 & 94.7 \\

%% file: tables/tab_bench_phia_split.tex
\code{llama-3.3-70b} & \textbf{55} & 38 & \textbf{55} & \textbf{52} & 41 & 49 & 43 & 50 & \textbf{100} \\
\code{llama-4-scout} & \textbf{56} & 51 & 54 & 52 & \textbf{54} & 48 & 43 & 43 & \textbf{93} \\
\code{mistral-small-3.1} & 40 & \textbf{48} & 45 & 38 & \textbf{50} & 43 & 42 & 17 & \textbf{58} \\
\code{gemma-4-26b} & 51 & \textbf{61} & 50 & 45 & \textbf{53} & 45 & 35 & 33 & \textbf{83} \\
\code{gpt-oss-20b} & \textbf{62} & 59 & 49 & \textbf{59} & 50 & 47 & 52 & 72 & \textbf{98} \\
\code{gpt-oss-120b} & \textbf{65} & 64 & 46 & \textbf{57} & 54 & 43 & 52 & 57 & \textbf{100} \\
\code{gemini-4-neon} & 43 & \textbf{50} & \textbf{50} & 42 & 45 & \textbf{47} & 53 & 53 & \textbf{65} \\
\code{gpt-6.2-astra} & 65 & \textbf{66} & 45 & 54 & \textbf{55} & 40 & 55 & 57 & \textbf{83} \\
\code{claude-fable-5.2} & \textbf{67} & 59 & 52 & \textbf{57} & 52 & 44 & 58 & 48 & \textbf{75} \\
\midrule
All models & \textbf{56} & 55 & 50 & \textbf{51} & \textbf{51} & 45 & 48 & 48 & \textbf{84} \\

%% file: tables/tab_bench_real.tex
\code{llama-3.3-70b} & 63 (55 to 70) & 49 (41 to 57) & \textbf{84} (77 to 89) \\
\code{llama-4-scout} & 66 (58 to 73) & 68 (60 to 75) & \textbf{82} (75 to 87) \\
\code{mistral-small-3.1} & 69 (61 to 76) & 77 (70 to 83) & \textbf{81} (74 to 86) \\
\code{gemma-4-26b} & 76 (68 to 82) & \textbf{89} (84 to 93) & 88 (81 to 92) \\
\code{gpt-oss-20b} & \textbf{91} (85 to 94) & \textbf{91} (86 to 95) & 80 (73 to 86) \\
\code{gpt-oss-120b} & 90 (84 to 94) & \textbf{93} (88 to 96) & 82 (75 to 87) \\

%% file: sections/path.tex
\subsection{Results along the path to the deployed broker}\label{sec:res:v2}

\begin{table}[t]
\centering\small
\caption{One path from the study to the broker as deployed, condition \CC{} on the same tasks at each point: the
study's \CC{}, \CC{}$_1$ after step~1 and \CC{}$_2$ after step~2 (\cref{sec:v2-design}). Accuracy in \%, except the
week-over-week question (correct runs) and privacy (leaks counted by the detector). $p$: exact McNemar, \CC{}$_2$
against the study's \CC{} (\CC{}$_2$-only / \CC{}-only). Fresh PHIA: the sample drawn for step~1, reused for
step~2, six open-weight models. Best per row in bold (fewest leaks for privacy).}
\label{tab:v3}
\setlength{\tabcolsep}{4pt}\footnotesize
\begin{tabular}{@{}l r c c c l@{}}
\toprule
 & & Study & After step 1 & After step 2 & \\
Suite & runs & \CC{} & \CC{}$_1$ & \CC{}$_2$ & $p$\\
\midrule
\input{tables/tab_v3.tex} 
\bottomrule
\end{tabular}
\end{table}

\begin{table}[t]
\centering\small
\caption{After step~1, on a fresh PHIA sample: on answerable questions as worded \sys{} was not detectably different
from raw context when pooled, better for two models and worse for two; refusal of unsupported questions
held. Six open-weight models, 120
answerable and 60 unsupported questions disjoint from the study's and from the planner samples; accuracy in \%. $p$:
exact McNemar, \CC{}$_1$ against \CA{} on answerable questions as worded (\CC{}$_1$-only / \CA{}-only). Best per row and group in bold.}
\label{tab:v2phia}
\setlength{\tabcolsep}{2.5pt}\footnotesize
\begin{tabular}{@{}l ccc ccc ccc c@{}}
\toprule
 & \multicolumn{3}{c}{Answerable, as worded} & \multicolumn{3}{c}{Answerable, PHIA labels} & \multicolumn{3}{c}{Unsupported: refused} & \\
\cmidrule(lr){2-4}\cmidrule(lr){5-7}\cmidrule(lr){8-10}
Agent model & \CA{} & \CB{} & \CC{}$_1$ & \CA{} & \CB{} & \CC{}$_1$ & \CA{} & \CB{} & \CC{}$_1$ & $p$\\
\midrule
\input{tables/tab_v2_phia.tex} 
\bottomrule
\end{tabular}
\end{table}

Per-question and per-cause breakdowns were added after the runs and were not pre-registered. \Cref{tab:v3} follows
\sys{} along the path of \cref{sec:v2-design}; these results come from one measurement of each step.

\paragraph{Step 1, H1: not detectably different from raw context when pooled, not for every model, and not on PHIA's
labels.} Over the six open-weight models, on questions the study never used, \CC{}$_1$ answered \vtwoWordedPooledC{}
\vtwoWordedPooledCIC{} of the answerable questions as worded, against \vtwoWordedPooledA{} \vtwoWordedPooledCIA{} for
raw context and \vtwoWordedPooledB{} \vtwoWordedPooledCIB{} for open tools (\cref{tab:v2phia}; paired, $p=$
\vtwoPooledMcn{}). Both pre-registered parts of H1 were met: \CC{}$_1$'s pooled
accuracy was above the study's \CC{} for the same six models (\vtwoStudyWordedC{}, where raw context scored
\vtwoStudyWordedA{}), and it moved from \vtwoGapStudy{} raw context to \vtwoGapNew{} it. The pooled figure
hides a split. \CC{}$_1$ was significantly more accurate than raw context for \code{llama-3.3-70b}
(\vtwoWordedLlamaThreeC{} against \vtwoWordedLlamaThreeA{}, $p=$ \vtwoMcnLlamaThree{}) and \code{llama-4-scout}
(\vtwoWordedLlamaScoutC{} against \vtwoWordedLlamaScoutA{}), not detectably different for \code{mistral-small-3.1} and
\code{gemma-4-26b}, and significantly less accurate for both \code{gpt-oss} models (\code{gpt-oss-120b}:
\vtwoWordedOssLargeC{} against \vtwoWordedOssLargeA{}). Against PHIA's own labels the gap grew: \CC{}$_1$ scored
\vtwoLabelPooledC{} against \vtwoLabelPooledA{} for raw context ($p$ \vtwoLabelPooledMcn{}), where the study's \CC{} had
scored \vtwoStudyLabelC{} against \vtwoStudyLabelA{}.

\paragraph{Step 1, what each change did.} Attributed against PHIA's labels as in \cref{sec:res:bench}, the
720 \CC{}$_1$ runs on answerable questions had 2 that received only a threshold flag, against 159 for the same models'
\CC{} runs in the study: counts now reach the agent model. Those counts follow the catalog's definitions, in which
sleep is time asleep, while PHIA's reference code counts time in bed; so threshold accuracy stayed low
(\vtwoThrAcc{} as worded for \CC{}$_1$, \vtwoThrAccA{} for raw context), and runs whose released value differed from the
label rose from 40 to 148, and those with PHIA's $N{+}1$-day window from 33 to 118, which is why the gap on labels
grew. The denial hint was carried by \vtwoDeniedWithAlt{} of the \vtwoDeniedRuns{} runs with a denied metric; in
\vtwoDeniedFollowed{} of them, about a quarter, the agent model asked again for the granted metric and received a
packet, and \vtwoDeniedFollowedCorrect{} of those were correct as worded.

\paragraph{Step 1, H2: met.} On the 60 unsupported questions, \CC{}$_1$ answered \code{unknown} in \vtwoUnsupPooledC{} of
runs, against \vtwoUnsupPooledA{} for raw context and \vtwoUnsupPooledB{} for open tools, above the pre-registered bar
of the study's overall rate (84\%; \vtwoStudyUnsupC{} for the same six models).

\paragraph{Step 1, H3: not met on two counts.} \emph{Synthetic questions:} \CC{}$_1$ was \vtwoRegAccuracyNew{} accurate
against \vtwoRegAccuracyOld{} for \CC{} on the same \vtwoRegAccuracyN{} runs (pooled $p=$ \vtwoRegAccuracyPooledP{};
\cref{tab:v2reg}), with one significant drop, \code{mistral-small-3.1}, and one
significant gain, \code{gpt-oss-120b}. By question, the loss is in the week-over-week comparison, right in
\vtwoRegQWeekNew{} of \vtwoRegQWeekN{} runs against \vtwoRegQWeekOld{}, because step~1's planner prompt more often
compared the two weeks before last; no other question moved by more than three runs. \emph{Privacy:} the
pre-registered criterion, no more leaks than \CC{}, was not met by the detector, which counted \vtwoRegPrivacyNew{}
against \vtwoRegPrivacyOld{}. Reading the four: two are false positives, a fill-in template and a request asking
the user for their figures; one restates a granted yes-or-no answer (``I slept under 6 hours''); and one states a
resting heart rate after the broker denied the whole request, so the number was invented. None follows a release
outside the grant. \emph{Real data} (reported, no criterion): \vtwoRegRealNew{} against \vtwoRegRealOld{}, no model
significantly different.

\paragraph{Step 2: the planner is given the comparison windows.}\label{sec:res:v3} Step~2 changes only the planner
prompt: for comparisons with the period before, it lists both windows for 3, 7, 14, 28 and 30 days and for this week
against last week, computed from the date as the prompt already did for single windows; it adds one comparison
example; and it refines step~1's compound-question rule, so that such a question is refused only when no part is
supported. A planner-alone check was run first, on two drafts of the prompt; its in-distribution requests come from
the synthetic week-over-week task, so that task's recovery is partly in-sample.

\paragraph{Step 2, the planner alone.} On a held-out set of 16 window questions with expected windows, written before
the change, the planner resolved the window correctly for 14 with the study's prompt, 12 with step~1's and 15 with
step~2's. On the 167 requests that agent models had sent for the week-over-week task, the number of comparison
plans shifted by a week was 2, 25 and 0 (13, 38 and 13 counting every operation), and the number of refusals 56, 32
and 28.

\paragraph{Step 2, synthetic questions: met.} \CC{}$_2$ was \vthreeAccThree{} accurate on the same \vthreeAccN{} runs,
against \vthreeAccOne{} for the study's \CC{} and \vthreeAccTwo{} for \CC{}$_1$ ($p=$ \vthreeAccP{} against \CC{};
\cref{tab:v3}); no model differed significantly from the study. On the week-over-week question it was right in
\vthreeWeekThree{} of \vthreeWeekN{} runs, against \vthreeWeekOne{} for \CC{} and \vthreeWeekTwo{} for \CC{}$_1$.

\paragraph{Step 2, fresh PHIA: refusal met; accuracy as worded not met.} On the answerable questions, as worded,
\CC{}$_2$ answered \vthreePhiaWordedThree{} against \vthreePhiaWordedTwo{} for \CC{}$_1$ ($p=$ \vthreePhiaWordedPvsTwo{})
and \vthreePhiaWordedA{} for raw context ($p=$ \vthreePhiaWordedP{}). The
pre-registered criterion, not lower than \CC{}$_1$, was not met: \CC{}$_2$ was one point lower (396 against 402 of 720),
a difference that is not significant. On PHIA's labels it scored \vthreePhiaLabelThree{} against
\vthreePhiaLabelTwo{} for \CC{}$_1$ and \vthreePhiaLabelA{} for raw context, so the gap on labels is unchanged. The
per-model split also stayed: both \code{llama} models were well above raw context (\code{llama-3.3-70b}
\vthreePhiaWordedLlamaThree{} against \vthreePhiaWordedLlamaThreeA{}), both \code{gpt-oss} models well below
(\code{gpt-oss-120b} \vthreePhiaWordedOssLarge{} against \vthreePhiaWordedOssLargeA{}). On unsupported questions
\CC{}$_2$ answered \code{unknown} in \vthreePhiaUnsupThree{} of runs, against \vthreePhiaUnsupTwo{} for \CC{}$_1$ and
\vthreePhiaUnsupA{} for raw context. The fresh PHIA sample had been used once before, in step~1, though not to write
this prompt.

\paragraph{Step 2, privacy: met against \CC{}$_1$, above the study.} On the \vthreePrivN{} runs shared with the study,
the detector counted \vthreePrivThree{} leaks, against \vthreePrivTwo{} for \CC{}$_1$ and \vthreePrivOne{} for the
study's \CC{}. Reading them: one is a fill-in message template with an example figure (``only getting about 4
hours''), one names a readiness score, a metric outside the grant, without a value and with none released, and one
states a resting heart rate after the broker denied the whole request, so the number was invented. None follows a
release outside the grant.

\paragraph{Step 2, real data.} On the open-weight models' real-data tasks \CC{}$_2$ was \vthreeRealThree{}
accurate, against \vthreeRealOne{} for the study's \CC{} ($p$ \vthreeRealP{}) and
\vthreeRealTwo{} for \CC{}$_1$, significantly higher than the study for \vthreeRealSig{} (uncorrected; all but
\code{gpt-oss-20b} after Holm correction over the six models). All 60 gains were on multi-day windows, mostly the
14-day resting heart rate (16 gained, none lost) and the 7-day step total (12 and none). The step-2 prompt was
written after the earlier real-data results were known, so this gain is not out-of-sample.
Against the study's raw context, \CC{}$_2$ was significantly more accurate for the four weaker models
(\code{llama-3.3-70b} \vthreeRealLlamaThree{} against \vthreeRealLlamaThreeA{}, $p$ \vthreeRealPALlamaThree{}) and no longer significantly less accurate for either \code{gpt-oss} model (\code{gpt-oss-120b} \vthreeRealOssLarge{}
against \vthreeRealOssLargeA{}, $p=$ \vthreeRealPAOssLarge{}; \code{gpt-oss-20b} \vthreeRealOssSmall{} against
\vthreeRealOssSmallA{}, $p=$ \vthreeRealPAOssSmall{}).

\paragraph{Where the path ends.} The broker as deployed is the one after step~2. Step~1 brought pooled accuracy on
fresh answerable PHIA questions, as worded, to within a point of raw context and let some denied runs
recover, but widened the gap on PHIA's labels, cost accuracy on one synthetic question and did not meet the privacy
criterion by the detector. Step~2 removed that synthetic regression and raised accuracy on real data (in-sample); it did
not move PHIA, where the strongest open models still did better with raw context.

%% file: tables/tab_v3.tex
Synthetic questions & 900 & \textbf{97} & 96 & \textbf{97} & 1.00 (14/15) \\
Week-over-week question (correct runs) & 90 & 83 & 73 & \textbf{88} & -- \\
Real data & 912 & 83 & 82 & \textbf{88} & $<$0.001 (60/9) \\
Privacy (leaks) & 300 & \textbf{1} & 4 & 3 & -- \\
Fresh PHIA, answerable, as worded & 720 & -- & \textbf{56} & 55 & -- \\
Fresh PHIA, answerable, PHIA labels & 720 & -- & \textbf{41} & 40 & -- \\
Fresh PHIA, unsupported refused & 360 & -- & 87 & \textbf{89} & -- \\

%% file: tables/tab_v2_phia.tex
\code{llama-3.3-70b} & 45 & 37 & \textbf{60} & \textbf{42} & 36 & 41 & 43 & 48 & \textbf{100} & 0.004 (27/9) \\
\code{llama-4-scout} & 57 & 47 & \textbf{74} & 49 & \textbf{56} & 52 & 45 & 40 & \textbf{92} & $<$0.001 (26/5) \\
\code{mistral-small-3.1} & 40 & \textbf{47} & 41 & 38 & \textbf{47} & 32 & 47 & 20 & \textbf{55} & 1.00 (19/18) \\
\code{gemma-4-26b} & 51 & \textbf{63} & 53 & 45 & \textbf{48} & 41 & 53 & 38 & \textbf{83} & 0.70 (15/12) \\
\code{gpt-oss-20b} & \textbf{68} & 64 & 52 & \textbf{54} & 47 & 40 & 60 & 78 & \textbf{98} & 0.003 (8/26) \\
\code{gpt-oss-120b} & \textbf{70} & 67 & 54 & \textbf{51} & 47 & 39 & 68 & 73 & \textbf{93} & $<$0.001 (5/24) \\
\midrule
All open-weight models & 55 & 54 & \textbf{56} & \textbf{47} & \textbf{47} & 41 & 53 & 50 & \textbf{87} & \\

%% file: sections/discussion.tex
\section{Discussion}\label{sec:discussion}

\subsection{What the results say}

\paragraph{Computing outside the agent model was associated with large gains where the agent model was weak.} The
accuracy gains in \cref{tab:accuracy} concentrate in models that were weak with raw context, and the per-question
results are consistent with an explanation: the questions those models got wrong in \CA{} and \CB{} need arithmetic
over many days, a fitted trend, or knowledge that two sources report the same steps, and a packet hands the agent
model a computed value with its caveats. The study does not isolate this cause. \CC{} bundles computed values,
data-quality caveats, a second, larger model as planner, and a tool description that tells the agent model to rely
on the numbers. The decomposition in \cref{sec:res:failures} shows that runs with a released packet were nearly all correct and that almost no run was left without a packet, which fits the explanation but is not an
ablation. The pattern agrees with
the program-aided reasoning literature~\citep{gao2022pal,chen2023pot} and with PHIA's finding that computing in code
beats reasoning over the numbers~\citep{merrill2024phia}; in \sys{} the agent model writes no code at all.

\paragraph{For models near ceiling, the benefit is not accuracy.} The strongest models answered almost every
synthetic question correctly in every condition, and we claim no accuracy gain for them there; the differences are at
most a few points. That is specific to the suite: on PHIA, as worded, \code{gpt-6.2-astra} and \code{claude-fable-5.2}
were less accurate with \sys{} (\phiaWordedAstraC{} against \phiaWordedAstraA{}, \phiaWordedFableC{} against
\phiaWordedFableA{}), and in the study both \code{gpt-oss} models, the strongest on real data, were less accurate
with it there (\cref{sec:res:bench}). Nor were the strongest models simply the best choice: with \sys{}, accuracy
depended less on the model, from \accCMin{} to \accCMax{} across all nine against \accAMin{} to \accAMax{} with raw
context, and high accuracy did not mean discretion. \code{gpt-oss-120b}, at \accOssLargeA{} with raw context, leaked
in \leakOssLargeA{} privacy runs with raw context and \leakOssLargeB{} with open tools, against \leakOssLargeC{} with
\sys{}, while the frontier models leaked in at most one of 20 runs in any condition (\cref{tab:privacymodels}). What \sys{} changes for these models is what they are given and what remains afterwards. The agent model
read \ctxRatioCAMin{} to \ctxRatioCAMax{} of its raw-context input; the release was limited to the level the question
needed, so a scheduling assistant receives a bucket rather than a night's sleep in minutes; and every release and
denial left a hash-chained receipt. For \code{claude-fable-5.2} and \code{gemini-4-neon} it also lowered the cost per
correct answer; for \code{gpt-6.2-astra} it raised it, and it added latency for every fast model.

\paragraph{Exposure and leakage are different quantities.} Exposure is fixed by the system before the agent model
runs, so it can be measured, audited and bounded; leakage depends on the model, the prompt and the adversary, and
prompt-level instructions reduce it without bounding it~\citep{li2026toolminimize,dong2026claw}. Our privacy results
show both sides. Where the request asked for data outside the role, raw context and open tools leaked often and
\sys{} did not, because the data was never delivered. Where the user asked for their own data to be used, as in the
manager message, the difference came from the release level: the \CC{} agent model received a bucket and had no
figure to restate. Upstream minimization has worked in other settings~\citep{bagdasarian2024airgap,abdelnabi2025firewalls,li2026toolminimize};
\sys{} applies it with computed release levels rather than a second model. Minimal answers can still compose into a
reconstruction of the record~\citep{dinur2003revealing,patil2025sumleaks}; the gate can cap releases per day, but
this study did not exercise that limit or run a reconstruction attack. Coarse levels limit values, not inference: a
bucket for temperature or HRV deviation can still hint at a cycle phase or an illness, and sensitive categories are
handled by catalog category, not by what a released value implies. Leak rates also depend on the detector; we
recommend that agent-privacy evaluations report detector precision and recall on a labeled sample and freeze the
detector before the runs it scores.

\paragraph{Where errors remain.} On the synthetic questions, \sys{}' remaining errors came almost all after a packet
was released: a partial total reported despite a caveat that it is incomplete, or \emph{yes} to a threshold when every
value received was above it.

\paragraph{What PHIA shows about the design.} PHIA exposed two decisions that cost accuracy on questions the broker
could answer. First, the gate releases the lowest level that can answer an \emph{operation}; for a threshold check
that is a yes-or-no flag, so a question asking \emph{how many} days met a condition received a flag although the
count was computed. The release level should follow the question asked, not only the operation. Second, a metric
outside the grant is denied without saying which granted metric could answer; the planner mapped ``heart rate'' to
the heart-rate metric, the grant held resting heart rate, and the agent model was left with a denial. A denial that
names an answerable alternative keeps the gate's guarantee and lets the agent model recover. Both changes are
general, not specific to PHIA, and form the first step of the path in \cref{sec:res:v2}: counts then reached the
agent model, a quarter of the denied runs recovered through the hint, and pooled accuracy on fresh answerable
questions, as worded, was not detectably different from raw context, though the gap on PHIA's labels grew. Two further observations are properties of the benchmark: some reference labels count time in bed or total
sleep where the question says sleep or light sleep, and ``last $N$ days'' is labeled as $N{+}1$ days. We report
PHIA as worded and with its labels for that reason. PHIA also shows the trade-off we chose: \sys{} refused
unsupported questions far more often than the baselines, and refused some answerable ones.

\paragraph{The open-tools condition depends on how the model calls tools.} Open tools were the worst condition for
some models and, on real data, the best for others. One model wrote most of its tool calls as text, which a harness that only
reads structured calls would have scored as unanswered; the harness executes such calls in every condition, and
with it the model's open-tools accuracy was \accLlamaScoutB{}. A comparison of MCP-style designs therefore depends on the serving stack as much as on the design,
and studies should state how text-format tool calls are handled.

\subsection{Implications for agent builders}

For people building agents over personal data, we draw four lessons. First, give agent models computed answers
with caveats rather than records; the packet contract cost the strongest models nothing measurable on the synthetic
questions, though they lost accuracy on PHIA and real data, and it helped the weaker ones. Second, decide disclosure in code, outside the model, and bind purpose at consent time rather
than in the request. Third, log both releases and denials with their reasons, which make planner and gate failures
traceable. Fourth, choose the release level from the question and make denials point to what the grant
does allow; a minimal answer that does not answer the question is not minimal, only unhelpful.

\subsection{Limitations of the design}

The gate cannot stop an agent model from misreading a packet, and a released packet is outside the system's
control once an agent's provider stores it. Within the ceiling the grant sets, what is requested is chosen by the
planner and the agent model, so the release is deterministic in its limits, not in its content. The operation set is
deliberately narrow, so \sys{} refuses questions a code-writing agent could answer (\phiaUnsupported{} of PHIA's
\phiaTotal{} objective questions). In the threat model the operator of the broker is trusted. So are the providers
hosting the planner, which sees the question text, and the agent model, which sees each packet. The gate governs the
flow to the agent, not the agent's onward flows.
Receipts are hash-chained but anchored on the same platform, so they detect tampering by others, not by the
operator, and a stolen agent token reaches only that agent's grant.

%% file: sections/scope.tex
\section{Scope of the Study}\label{sec:scope}

\paragraph{Synthetic cohort.} The accuracy and privacy studies use ten synthetic people whose ground truth is known
by construction. This lets us score every answer against what actually happened rather than against what a device
or the broker reports, and lets us place each data-quality trap deliberately. The cost is realism and size: real
exports have more varied gaps, devices and artifacts than ten seeded profiles, each person carries one trap, each
task was run once per model and condition, and the questions name their windows explicitly. Tasks are clustered
within people, which is why we report a bootstrap over people next to the per-task tests; with ten people as
clusters its percentile intervals may be narrow, and the models are treated as fixed. The results describe these
traps and questions, not the distribution of questions real users ask; the real-data suite (\cref{sec:res:bench})
addresses part of this.

\paragraph{Scoring and broker designed by the same authors.} The reference rules and the broker were written by the
same team, and some definitions coincide. The HRV reference is \emph{flat} for every person without an HRV trap,
while the broker calls a trend only when the slope's $t$ statistic reaches 2.5. ``Sleeping less than usual'' is
defined by quartiles of the prior nights, and the broker compares with an interquartile baseline; \usualLenient{} of
the \usualTasks{} such tasks accepts either answer. The planner prompt resolves the windows the questions use
(``ending yesterday''). Each of these favors \CC{} to an unknown degree; the PHIA and real-data suites, whose
references the broker's authors did not define or computed independently, are the check.

\paragraph{Public benchmark.} The planner and compute-engine studies use PHIA's public objective question set and
its released data (commit \code{\phiaCommit}), under its CC BY-NC 4.0 license. The planner was chosen on the
selection sample, its refusal list follows PHIA's unsupported categories, and the directly answerable subsets are
small (\plLlamaThreeNAns{} and \plHeldLlamaThreeNAns{} questions), so agreement figures carry wide uncertainty. The
compute-engine study uses hand-mapped queries and covers the \phiaMapped{} questions inside the operation set.

\paragraph{Models and serving.} We evaluated the \nModels{} models listed in \cref{sec:models}, each through one
provider and with provider-default sampling. Open-weight models served elsewhere, with other quantization or
batching, may behave differently~\citep{yuan2025nondeterminism,he2025nondeterminism}. All runs share one evaluation
date. Privacy runs for the frontier models use \privPeopleOther{} of the \privPeopleOpen{} synthetic people. Repeatability covers two
sleep questions and the open-weight models only.

\paragraph{Leak scoring.} Leaks are scored by deterministic rules over the agent model's final text, with no model as
a judge, corrected after a check against one annotator's labels and not re-validated on a fresh labeled sample
(\cref{sec:res:privacy}).

\paragraph{Baselines and ablations.} \CA{} and \CB{} represent the two prevailing designs, not every possible one. We
did not compare against PHIA's code-writing agent, a model-mediated minimizer~\citep{bagdasarian2024airgap}, a
deterministic tool gate without the metric layer~\citep{shi2025progent}, \CC{} with an oracle planner, or \CA{} with
self-consistency at matched compute~\citep{wang2023selfconsistency}; each would separate a different part of
\sys{}' effect.

%% file: sections/conclusion.tex
\section{Conclusion}\label{sec:conclusion}

We measured what changes when an AI agent asks about wearable data through a broker that computes the answer,
releases only what the question needs, and records every decision. Holding the agent and planner models fixed,
\sys{} raised pooled accuracy over raw context by \bootPooledCAEst{} points (\bootPooledCACI{}), about
\sensCapHrvEst{} excluding raw-context runs that hit the output limit and a trend question whose reference matches the
broker's rule; accuracy rose significantly for \nSigCA{} of \nTestedCA{} models (\sensCapNSig{} excluding those runs),
fell for none, and depended less on the choice of model. Raw context put \expRatioAC{} times as much of the record in
front of the model (for a 45-day export), and \sys{} leaked in \leakPooledC{} of privacy runs against \leakPooledA{}
for raw context and \leakPooledB{} for open tools; in every scenario the grant or release level withheld each scored
quantity, so its one leak was an invented figure. On PHIA, \sys{} refused out-of-scope questions far more often than
either baseline but answered fewer answerable ones; after two pre-registered steps, the deployed broker matched raw
context on fresh answerable questions as worded (\vthreePhiaWordedThree{} against \vthreePhiaWordedA{}) but not on
PHIA's labels (\vthreePhiaLabelThree{} against \vthreePhiaLabelA{}), and the strongest open-weight models still did
better with raw context there. Code, data and the scripts that produce every number in this paper are available at
\codeurl{}.

%% file: sections/ethics.tex
\phantomsection
\begingroup\sloppy
\section*{Ethics and Reproducibility Statement}\label{sec:ethics}
\addcontentsline{toc}{section}{Ethics and Reproducibility Statement}

The study used synthetic data and the public PHIA, PMData and LifeSnaps datasets under their licenses. The real-data
analyses are secondary analyses of PMData and LifeSnaps, which their authors released publicly without direct
identifiers; using them is not human-subjects research as defined in 45 CFR 46.102(e); no re-identification was
attempted and no new data were collected. Question text and reference
answers from PHIA are not redistributed; the repository stores question identifiers and hashes only. The grants in
the privacy scenarios are illustrative, chosen by the authors. \sys{} answers wellness questions about a person's own
data and makes no medical claims: its packets carry a flag stating that they are not medical advice, the planner
refuses requests for grades, diagnoses or medical advice, and nothing in this paper is a clinical claim.

The broker, the synthetic generator, the question set and scoring rules, the privacy scenarios, the agent prompts,
the leak-detector audit with its hand labels, and the per-run records are available at \codeurl{}. Each record holds
the final text, parsed answer, tool calls, packet levels and gate reasons; records of the synthetic, real-data and
repeatability suites also hold every packet released. The pre-registration of the two steps after the study is in
the repository, which maps every table, figure and number to the script that produces it.
\par\endgroup

%% file: sections/appendix.tex
\section{Additional Tables and Figures}\label[appendix]{app:tables}

\begin{table}[h]
\centering\small
\caption{Accuracy first, then what each answer costs. Bold marks the best value per model: the highest accuracy and the lowest cost; where the
95\% accuracy intervals overlap, the difference is not significant (see the McNemar $p$ values in \cref{tab:accuracy}).
Context = median input tokens per question read by the agent model, and in \CC{} separately by the broker's planner
(\code{llama-3.3-70b}); output = median output tokens of the agent model. Cost is the mean USD per question
(provider-reported for frontier models, Workers AI list prices for open-weight models, planner included in \CC{});
cost per correct answer is total cost divided by correct answers.}
\label{tab:usage}
\setlength{\tabcolsep}{3pt}\footnotesize
\begin{tabular}{@{}l c c r r r r r@{}}
\toprule
 & & & \multicolumn{2}{c}{Context (input tokens)} & & \multicolumn{2}{c}{Cost (USD)}\\
\cmidrule(lr){4-5}\cmidrule(lr){7-8}
Agent model & Cond. & Accuracy, \% (95\% interval) & Agent & Planner & Output & per question & per correct\\
\midrule
\input{tables/tab_usage.tex} 
\bottomrule
\end{tabular}
\end{table}

\begin{figure}[h]
\centering
\includegraphics[width=\textwidth]{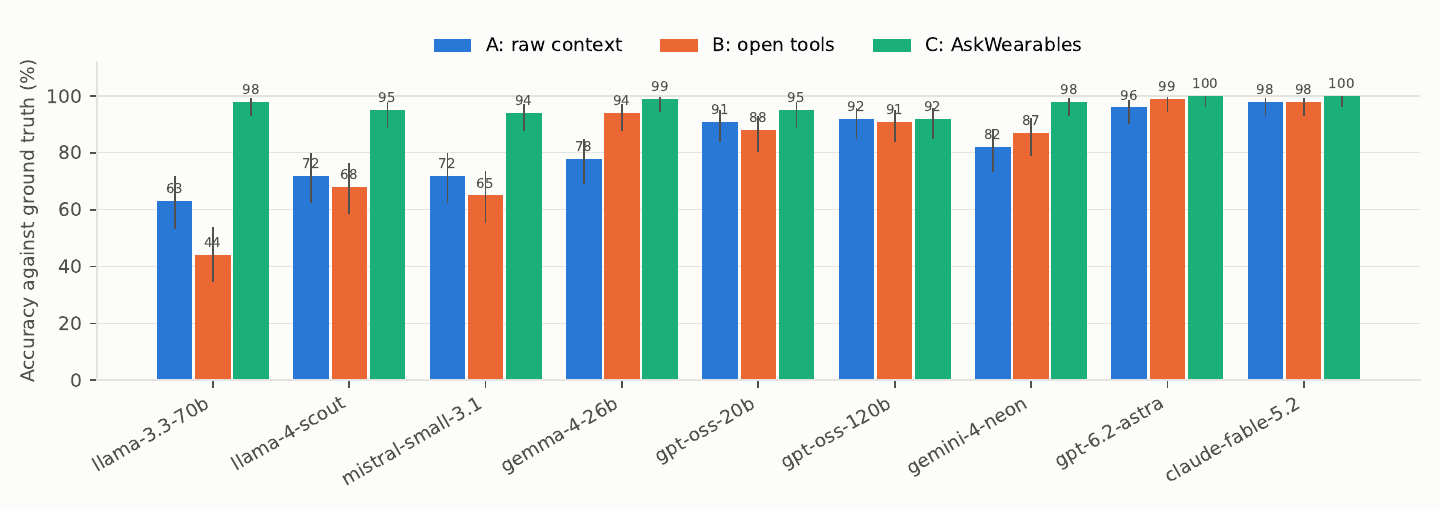}
\caption{Accuracy against ground truth by model and condition, with Wilson 95\% intervals (the data of
\cref{tab:accuracy}).}
\label{fig:accuracy}
\end{figure}

\begin{table}[h]
\centering\small
\caption{Step~1 against the study's \sys{} on the same tasks, per model (condition \CC{} only): the synthetic
questions, and real data for the open-weight models. Accuracy in \%; \CC{}$_1$-only and \CC{}-only count tasks
right only after step~1 or only in the study; $p$ is the exact McNemar test. Best per row in bold.}
\label{tab:v2reg}
\setlength{\tabcolsep}{4pt}\footnotesize
\begin{tabular}{@{}l r c c r r c@{}}
\toprule
Agent model & tasks & \CC{} & \CC{}$_1$ & \CC{}$_1$-only & \CC{}-only & $p$\\
\midrule
\multicolumn{7}{@{}l}{\emph{Synthetic questions}}\\
\input{tables/tab_v2_accuracy.tex} 
\midrule
\multicolumn{7}{@{}l}{\emph{Real data}}\\
\input{tables/tab_v2_real.tex} 
\bottomrule
\end{tabular}
\end{table}

\section{Reproducibility Notes}\label[appendix]{app:repro}

\paragraph{Harness.} Tool-call arguments that are not valid JSON are passed to the tool as an empty object, and the
tool result says so. The planner in the study was called without schema-constrained decoding, and its JSON was
validated afterwards.

\paragraph{Deployed broker.} Besides the two steps of \cref{sec:res:v2}, the deployed broker requests
schema-constrained decoding from the model provider, and it drops the keyword pre-check for sensitive categories:
the planner prompt and the code check on the plan refuse them (all 20 sensitive test questions and none of 10
near-misses were refused without the pre-check).

%% file: tables/tab_usage.tex
\code{llama-3.3-70b} & A & 63 (53 to 72) & 4{,}908 & -- & 200 & 0.00194 & 0.00308 \\
 & B & 44 (35 to 54) & 1{,}375 & -- & 52 & \textbf{0.00080} & 0.00182 \\
 & C & \textbf{98} (93 to 99) & 1{,}039 & 1{,}860 & 93 & 0.00107 & \textbf{0.00109} \\
\midrule
\code{llama-4-scout} & A & 72 (63 to 80) & 4{,}882 & -- & 308 & 0.00162 & 0.00225 \\
 & B & 68 (58 to 76) & 1{,}289 & -- & 217 & \textbf{0.00061} & \textbf{0.00089} \\
 & C & \textbf{95} (89 to 98) & 950 & 1{,}860 & 81 & 0.00095 & 0.00100 \\
\midrule
\code{mistral-small-3.1} & A & 72 (63 to 80) & 6{,}382 & -- & 284 & 0.00242 & 0.00337 \\
 & B & 65 (55 to 74) & 1{,}189 & -- & 86 & \textbf{0.00059} & \textbf{0.00090} \\
 & C & \textbf{94} (88 to 97) & 1{,}028 & 1{,}861 & 97 & 0.00113 & 0.00120 \\
\midrule
\code{gemma-4-26b} & A & 78 (69 to 85) & 6{,}510 & -- & 1{,}568 & 0.00124 & 0.00158 \\
 & B & 94 (88 to 97) & 1{,}172 & -- & 979 & \textbf{0.00054} & \textbf{0.00057} \\
 & C & \textbf{99} (95 to 100) & 888 & 1{,}864 & 430 & 0.00091 & 0.00092 \\
\midrule
\code{gpt-oss-20b} & A & 91 (84 to 95) & 4{,}896 & -- & 381 & 0.00115 & 0.00126 \\
 & B & 88 (80 to 93) & 953 & -- & 376 & \textbf{0.00038} & \textbf{0.00043} \\
 & C & \textbf{95} (89 to 98) & 910 & 1{,}858 & 253 & 0.00092 & 0.00097 \\
\midrule
\code{gpt-oss-120b} & A & \textbf{92} (85 to 96) & 4{,}890 & -- & 470 & 0.00218 & 0.00237 \\
 & B & 91 (84 to 95) & 957 & -- & 312 & \textbf{0.00074} & \textbf{0.00082} \\
 & C & \textbf{92} (85 to 96) & 951 & 1{,}862 & 243 & 0.00132 & 0.00143 \\
\midrule
\code{gemini-4-neon} & A & 82 (73 to 88) & 6{,}530 & -- & 1{,}578 & 0.04811 & 0.05867 \\
 & B & 87 (79 to 92) & 1{,}584 & -- & 1{,}285 & 0.04674 & 0.05372 \\
 & C & \textbf{98} (93 to 99) & 1{,}126 & 1{,}876 & 618 & \textbf{0.01963} & \textbf{0.02003} \\
\midrule
\code{gpt-6.2-astra} & A & 96 (90 to 98) & 4{,}833 & -- & 74 & \textbf{0.01162} & \textbf{0.01211} \\
 & B & 99 (95 to 100) & 800 & -- & 108 & 0.01299 & 0.01312 \\
 & C & \textbf{100} (96 to 100) & 917 & 1{,}882 & 140 & 0.01342 & 0.01342 \\
\midrule
\code{claude-fable-5.2} & A & 98 (93 to 99) & 7{,}351 & -- & 354 & 0.07725 & 0.07882 \\
 & B & 98 (93 to 99) & 2{,}255 & -- & 516 & 0.04489 & 0.04581 \\
 & C & \textbf{100} (96 to 100) & 1{,}771 & 1{,}875 & 334 & \textbf{0.03028} & \textbf{0.03028} \\

%% file: tables/tab_v2_accuracy.tex
\code{llama-3.3-70b} & 100 & \textbf{98} & 94 & 0 & 4 & 0.12 \\
\code{llama-4-scout} & 100 & \textbf{95} & 92 & 0 & 3 & 0.25 \\
\code{mistral-small-3.1} & 100 & \textbf{94} & 88 & 0 & 6 & 0.03 \\
\code{gemma-4-26b} & 100 & \textbf{99} & 98 & 0 & 1 & 1.00 \\
\code{gpt-oss-20b} & 100 & \textbf{95} & 94 & 2 & 3 & 1.00 \\
\code{gpt-oss-120b} & 100 & 92 & \textbf{99} & 7 & 0 & 0.02 \\
\code{gemini-4-neon} & 100 & \textbf{98} & 97 & 1 & 2 & 1.00 \\
\code{gpt-6.2-astra} & 100 & 100 & 100 & 0 & 0 & 1.00 \\
\code{claude-fable-5.2} & 100 & 100 & 100 & 0 & 0 & 1.00 \\
\midrule
All & 900 & \textbf{97} & 96 & & & \\

%% file: tables/tab_v2_real.tex
\code{llama-3.3-70b} & 152 & \textbf{84} & 82 & 3 & 6 & 0.51 \\
\code{llama-4-scout} & 152 & \textbf{82} & 80 & 4 & 7 & 0.55 \\
\code{mistral-small-3.1} & 152 & 81 & \textbf{82} & 6 & 4 & 0.75 \\
\code{gemma-4-26b} & 152 & \textbf{88} & 85 & 3 & 7 & 0.34 \\
\code{gpt-oss-20b} & 152 & 80 & 80 & 5 & 6 & 1.00 \\
\code{gpt-oss-120b} & 152 & 82 & \textbf{86} & 14 & 7 & 0.19 \\
\midrule
All & 912 & \textbf{83} & 82 & & & \\